%% file: template.tex
\documentclass{article}

\usepackage{arxiv}

\usepackage[utf8]{inputenc} 
\usepackage[T1]{fontenc}    
\usepackage{hyperref}       
\usepackage{url}            
\usepackage{booktabs}       
\usepackage{multirow}
\usepackage{makecell}
\usepackage{amsfonts}       
\usepackage{nicefrac}       
\usepackage{amsmath}
\usepackage{microtype}      
\usepackage{cleveref}       
\usepackage{lipsum}         
\usepackage{algorithm}
\usepackage{algorithmic}
\usepackage{longtable}

\usepackage{subcaption}
\usepackage{graphicx}
\usepackage{natbib}
\usepackage{doi}

\usepackage{tikz}
\usepackage{graphicx}
\usetikzlibrary{shapes.geometric, shapes.misc, arrows.meta, positioning, fit, backgrounds, calc, matrix, decorations.pathreplacing}
\usepackage{helvet}
\usepackage{amsmath}
\usepackage{amssymb}

\definecolor{primary}{RGB}{41, 128, 185}     
\definecolor{secondary}{RGB}{39, 174, 96}    
\definecolor{accent}{RGB}{211, 84, 0}        
\definecolor{darktext}{RGB}{44, 62, 80}      
\definecolor{lightbg}{RGB}{248, 249, 250}    
\definecolor{bordergray}{RGB}{189, 195, 199} 
\definecolor{rulebg}{RGB}{253, 246, 227}     

\title{Neural-Symbolic Knowledge Tracing: Injecting Educational Knowledge into Deep Learning for Responsible Learner Modelling}

\newif\ifuniqueAffiliation
\uniqueAffiliationtrue

\ifuniqueAffiliation 
\author{ 
 Danial Hooshyar \\
  School of Digital Technologies\\
  Tallinn University, Estonia\\
  Faculty of Information Technology\\
  Faculty of Education and Psychology\\
  University of Jyv\"askyl\"a\\
  Finland\\
  \texttt{danial.hooshyar@tlu.ee} \\
   \And
 Gustav \v{S}\'{\i}r\\
  Department of Computer Science\\
  Czech Technical University\\
  Czech Republic\\
  \texttt{gustav.sir@cvut.cz} \\
    \And
 Yeongwook Yang \\
  Department of Computer Science and Engineering\\
  Kangwon National University\\
  Republic of Korea \\
  \texttt{yeongwook.yang@kangwon.ac.kr} \\
    \And
Tommi K\"arkk\"ainen\\
  Faculty of Information Technology\\
  University of Jyv\"askyl\"a\\
  Finland\\
  \texttt{tommi.karkkainen@jyu.fi} \\
    \And  
 Raija H\"am\"al\"ainen\\
  Faculty of Education and Psychology\\
  University of Jyv\"askyl\"a\\
  Finland\\
  \texttt{raija.h.hamalainen@jyu.fi} \\
    \And
 Ekaterina Krivich \\
  School of Digital Technologies\\
  Tallinn University\\
  Estonia\\
  \texttt{ekaterina.krivich@tlu.ee} \\
    \And
Mutlu Cukurova \\
  UCL Knowledge Lab \\
  UCL Centre for Artificial Intelligence\\
  University College London\\
  UK \\
  \texttt{m.cukurova@ucl.ac.uk} \\
    \And
 Dragan Ga\v{s}evi\'{c}\\
  Faculty of Education\\
  School of Computing and Data Science\\
  The University of Hong Kong\\
  Centre for Learning Analytics\\
  Monash University\\
  Australia\\
  \texttt{dragan.gasevic@monash.edu} \\
    \And
 Roger Azevedo \\
  School of Modeling Simulation and Training\\
  University of Central Florida\\
  US\\
  \texttt{roger.azevedo@ucf.edu} \\
}
\else
\usepackage{authblk}

\newbox{\orcid}\sbox{\orcid}{\includegraphics[scale=0.06]{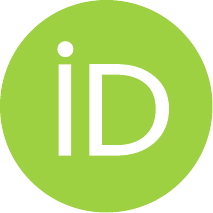}} 
\author[1]{%
	\href{https://orcid.org/0000-0000-0000-0000}{\usebox{\orcid}\hspace{1mm}David S.~Hippocampus\thanks{\texttt{hippo@cs.cranberry-lemon.edu}}}%
}
\author[1,2]{%
	\href{https://orcid.org/0000-0000-0000-0000}{\usebox{\orcid}\hspace{1mm}Elias D.~Striatum\thanks{\texttt{stariate@ee.mount-sheikh.edu}}}%
}
\affil[1]{Department of Computer Science, Cranberry-Lemon University, Pittsburgh, PA 15213}
\affil[2]{Department of Electrical Engineering, Mount-Sheikh University, Santa Narimana, Levand}
\fi

\renewcommand{\shorttitle}{\textit{arXiv} Template}

\hypersetup{
pdftitle={A template for the arxiv style},
pdfsubject={q-bio.NC, q-bio.QM},
pdfauthor={David S.~Hippocampus, Elias D.~Striatum},
pdfkeywords={First keyword, Second keyword, More},
}

\begin{document}
\maketitle
\clearpage   

\begin{abstract}
The growing adoption of artificial intelligence (AI) in education, particularly large language models (LLMs), has increased attention to intelligent tutoring systems. However, recent research show that LLMs alone often exhibit shallow adaptivity and struggle to reliably model learners’ evolving knowledge over time. This limitation highlights the need for dedicated learner modelling approaches that explicitly track knowledge progression. While deep knowledge tracing approaches have shown strong predictive performance for learner modelling tasks, their opaque nature and susceptibility to biases or spurious correlations in data can hinder alignment with pedagogical principles. In response, this study proposes a novel neural-symbolic deep knowledge tracing approach, called \texttt{Responsible-DKT}, which integrates symbolic educational knowledge (e.g., rules representing \textit{mastery} and \textit{non-mastery} states) into sequential neural models for responsible learner modelling. This paper reports the findings of experiments on a real-world dataset of 6th-grade students’ interactions in Maths learning, collected in September 2021, showing that \texttt{Responsible-DKT} outperforms both a neural-symbolic baseline without knowledge injection and a fully data-driven \texttt{PyTorch} implementation of DKT in predictive accuracy across different training ratios and sequence lengths. The model achieves over 0.80 AUC with only 10\% of training data and reaches up to 0.90 AUC, improving performance by up to 13\% over the baseline models. It also demonstrates improved temporal reliability, producing lower early- and mid-sequence prediction errors and the lowest prediction inconsistency rates across sequence lengths, indicating that prediction updates remain directionally aligned with observed student responses over time. Furthermore, the neural-symbolic architecture provides intrinsic interpretability through its grounded computation graph, which explicitly reveals the decision-making logic behind each prediction, enabling both local and global explanations. This transparency also allows pedagogical assumptions embedded in the rules to be empirically evaluated, showing that patterns of repeated incorrect responses (\textit{non-mastery}) play a stronger role in prediction updates. These findings suggest that hybrid human–AI approaches such as neural-symbolic computing improve predictive performance and interpretability, address data limitations, and support human-centred AI by bridging educational knowledge with advanced machine learning methods, enabling more responsible AI designs in educational contexts and beyond. 
\end{abstract}


\section{Introduction}
\label{sec:introduction}
Artificial intelligence (AI) technologies, such intelligent tutoring systems (ITSs), are increasingly being integrated into educational settings, with large language models (LLMs) emerging as one of the most widely used AI technologies \citep{azevedo2023theories,yan2024promises}. Studies demonstrate their potential to enhance student engagement in programming \citep{kumar2023impact, lyu2024evaluating}, strengthen writing skills \citep{benvenuti2023artificial}, support teachers through automated tutoring and grading \citep{labadze2023role, miroyan2025analyzing}, among other applications. Their ability to generate contextually relevant responses and immediate feedback has increased their appeal in both classrooms and educational research \citep{kaliisa2026does}. Governments are increasingly moving toward formalizing the role of AI within educational systems. For instance, Estonia's AI Leap initiative\footnote{\url{https://tihupe.ee/en/}} requires the integration of AI technologies across high schools. Meanwhile, emerging regulatory policies classify AI applications in education—similar to those in healthcare—as \textit{high-risk} \citep{europeanunion2024artificial, saarela2025eu}. This has strengthened calls for responsible AI guided by principles such as fairness, transparency, accountability, and human agency \citep{arrieta2020explainable, eitel2020beyond, hooshyar2025towards, pargman2024towards}. However, while \textit{responsible use of AI in education} is widely discussed, far less attention has been given to \textbf{how these systems are designed and developed to enable responsible use}.

Meanwhile, the individualized response generation of LLMs has fostered a \textbf{misconception} that they can function as general-purpose engines for adaptive education. Recent studies have attempted to develop "LLM tutors" through knowledge augmentation, fine-tuning, etc \citep{maurya2025unifying, molina2024leveraging, wang2025llm}. However, effectiveness of such LLM Tutors often under-performs that of well-established ITSs \citep {gasevic2026generative} and some studies report even near null or negative effects on learning gains \citep{bastani2025generative,eames2026computer}. such approaches often overlook the need to accurately assess learners' evolving knowledge and generate stable, evidence-based adaptive decisions \citep{corbett1994knowledge, hooshyar2025problems, koedinger2023astonishing,duboulay2023handbook}. For decades, adaptive educational technologies have addressed this challenge through learner modelling. ITSs, for example, rely on computational models that represent learners' knowledge, skills, misconceptions, progress, or affective states \citep{abyaa2019learner}. Grounded in theories of cognition and pedagogy, these models analyse learner–system interactions to monitor learning and adapt instruction accordingly \citep{azevedo2023theories, conati2023student, krivich2025systematic}, enabling transparent and pedagogically meaningful adaptivity.  
By contrast, many tutoring approaches based on LLMs only lack such explicit and theory-grounded mechanisms for representing learners’ knowledge over time, limiting their capacity to evaluate and adapt to individual learners’ evolving needs \citep{gasevic2026generative}. This limitation is also aligned with broader concerns about the reliability and educational validity of LLM-based systems \citep{yan2024promises}. Recent research has documented several limitations of LLMs, such as the generation of inaccurate content (hallucinations) \citep{qian2026towards}, unreliable reasoning behaviour \citep{zhao2025chain}, and non-transparent decision mechanisms \citep{zhao2023survey}. In educational contexts, these issues may manifest in subtle but consequential ways. For example, generative AI models may reproduce harmful or biased assumptions in instructional feedback \citep{du2025benchmarking, hooshyar2025towards}. Behavioural analyses of LLM-based assessment also reveal systematic limitations in how language models evaluate student writing, especially when dealing with negation structures and multilingual content \citep{karumbaiah2024evaluating}.

Most importantly, recent empirical studies question whether LLMs can support the core function of adaptive educational systems: 
model and track student knowledge, skills, and learning over time in a reliable, accurate and tracable manner. For instance, in a controlled empirical comparison, \citet{hooshyar2025problems} evaluated an open-source LLM, with and without task-specific fine-tuning, against a deep knowledge tracing model—amongst the most successful learner-modelling methods—using a large-scale open dataset. Their evaluation combined standard predictive metrics with analyses examining sequential stability and temporal coherence. Although the LLM improved after extensive fine-tuning, it still failed to match the performance of DKT and produced unstable mastery estimates, incorrect directional updates, and higher prediction errors during early learning interactions. Complementary evidence is reported by \citet{borchers2025can}, who investigated how LLMs adapt tutoring responses when critical learner-context information is removed. Their results showed that LLM outputs changed only minimally across altered contexts, indicating limited sensitivity to the learner's state. Collectively, these findings suggest that despite their potentials, LLM-based tools struggle to produce the stable, evidence-based adaptive decisions required for reliable learner modelling. One key reason for this limitation is that LLMs do not construct explicit learner models for each learner representing their evolving knowledge and skills (e.g., domain-specific content knowledge and metacognitive skills) over time. Without such representations, they may not reliably track mastery, estimate learning trajectories, or model the dynamics of skill acquisition. In contrast, (predictive) learner modelling approaches analyse sequences of student interactions to estimate mastery probabilities and learning progress.

Beyond empirical limitations, \textbf{responsible use of AI systems presupposes responsible design and development}. Technical adjustments, e.g., fine-tuning LLMs to minimise hallucinations or reduce bias, may tackle surface-level problems, yet they fail to address more fundamental structural limitations \citep{hooshyar2025towards, lee2024life, resnik2024large}. LLMs remain opaque "black-box" systems whose internal reasoning processes are difficult to interpret \citep{hooshyar2025towards, singh2024rethinking}. As a result, their generated explanations should not be mistaken for faithful representations of how decisions are made. These limitations collectively highlight the continuing importance of learner modelling in adaptive educational systems and suggest that, rather than replacing it, LLMs are more appropriately deployed as pedagogical interfaces (e.g., conversational tutors that explain concepts) or content generators paired with dedicated learner-modelling components to ensure responsible and pedagogically meaningful support. Understanding how such learner modelling systems can be designed in accordance with responsible AI principles therefore becomes a critical research challenge \citep{goellner2024responsible,hooshyar2025towards}. 

\subsection{Learner modelling in AI for education}
\label{subsec:learner_modelling}

AI has been used in education for two broad purposes: enhancing practice and advancing research. In practice, it supports personalized instruction, automated feedback, scalable learner support, and administrative tasks such as identifying students at risk of failure or dropout \citep{alwarthan2022explainable, qin2023early}. In research, AI supports the discovery of pedagogical insights by analysing educational data through data mining approaches \citep{baker2016educational, hooshyar2019mining, hooshyar2025towardsa}. A key component supporting both of these purposes is learner modelling, also often referred to as student modelling, which aims to represent learners' cognitive and non-cognitive attributes in a structured manner \citep{abyaa2019learner, conati2023student, duboulay2023handbook}. By analysing learners' interactions with systems, these models estimate knowledge states and learning progress, enabling real-time assessment and personalized support.

Broadly, learner modelling approaches fall into two methodological traditions: \textit{symbolic} and \textit{sub-symbolic}, each offering strengths and limitations \citep{holmes2022artificial, hooshyar2025towards}. Symbolic approaches emphasise transparency and interpretability, often representing knowledge through explicit rules or probabilistic structures. However, these methods frequently face challenges when dealing with uncertainty, complex learning processes, or scalability \citep{ilkou2020symbolic}. Within the learner modelling domain, classical knowledge tracing methods exemplify this tradition \citep{abdelrahman2023knowledge}. Bayesian Knowledge Tracing (BKT), for example, models learning as a Hidden Markov process with binary latent states indicating whether a skill is mastered or not. The model often relies on parameters such as prior knowledge, learning rate, guess probability, and slip rate \citep{daly2011learning, mao2018deep}. Because these parameters are interpretable and grounded in cognitive theory, BKT has been widely adopted in educational applications. However, its binary mastery assumption oversimplifies the continuous nature of learning. Moreover, the standard formulation does not account for forgetting or interactions among multiple skills, and parameter estimation problems may lead to suboptimal model solutions \citep{pelanek2017bayesian, saric2024twenty}. To address some of these limitations, Performance Factor Analysis (PFA) was introduced by \citet{pavlik2009performance}. Instead of relying on latent mastery states, PFA employs logistic regression to estimate the likelihood of a correct response by considering a learner's accumulated history of successes and failures. This allows for more fine-grained skill modelling and supports the representation of multiple skills simultaneously. Despite these advantages, PFA still depends on manually engineered features as predictors of learners' states and remains limited in its ability to capture complex temporal dependencies or learner-specific trajectories \citep{abdelrahman2023knowledge, gong2011construct, pavlik2021logistic}.

Sub-symbolic methods, e.g., deep neural networks, frequently perform better than symbolic and are particularly effective at handling complex (non)sequential data \citep{gervet2020deep}. Unlike symbolic models, these methods learn representations directly from (sequential) interaction data. This reduces the need for extensive manual feature engineering, while often leading to higher predictive accuracy \citep{garcez2023neurosymbolic, hooshyar2025towards, ilkou2020symbolic}. A widely recognised example of this paradigm is Deep Knowledge Tracing (DKT) \citep{piech2015deep}, which employs recurrent neural networks to capture the temporal evolution of students' knowledge. DKT marked a major methodological shift in learner modelling, moving from interpretable but rigid symbolic approaches to flexible, data-driven sub-symbolic models. Since its introduction, the field has seen an extensive wave of research aimed at refining DKT and improving prediction accuracy. As shown in the systematic review of DKT conducted by \citet{krivich2025systematic}, the literature has progressed from early recurrent neural networks-based models \citep{piech2015deep}, to memory-augmented approaches \citep[e.g.,][]{zhang2017dynamic}, to graph-based models \citep[e.g.,][]{yang2020gikt}, and most recently to hybrid methods that integrate multiple paradigms \citep[e.g.,][]{abdelrahman2022deep}.

Despite recent advances, DKT inherits many of the challenges common to deep neural networks and faces methodological challenges that reduce its practical value and adoption in practice \citep{wang2023wrong, yeung2018addressing}. First, in contrast to symbolic knowledge tracing approaches, DKT operates primarily as a fully data-driven model and therefore cannot directly incorporate structured educational knowledge. This includes elements such as causal relationships, theoretical models of learning, the role of specific variables, or interactions that affect learning outcomes—as well as knowledge provided by educators and domain experts—limiting both pedagogical alignment and interpretability \citep{hooshyar2025towards}. The absence of embedded domain knowledge also constrains educators' ability to engage with the model and ensure consistency with instructional goals \citep{celik2022promises}. Second, relying primarily on raw data, unlike symbolic approaches such as BKT that incorporate predefined structure and assumptions, increases the likelihood that model predictions are influenced by spurious patterns or hidden biases. Multiple studies have shown that deep learning–based (KT) models are highly sensitive to data quality issues (e.g. noisy responses and imbalanced distributions), sparsity, and other irregularities—which can lead to misleading inferences \citep{baker2022algorithmic, hooshyar2024augmenting, hooshyar2025towardsa, tato2022infusing}. For instance, \citet{cui2024model} showed that widely used DKT benchmarks such as ASSIST09, ASSIST17, and EdNet suffer from severe class imbalance. When evaluated on balanced datasets, model performance dropped substantially, suggesting that predictions were partly driven by answer-distribution biases rather than accurate representations of student knowledge. Third, predictions can sometimes behave inconsistently over time (i.e., issues of sequential stability of predictions \citep{krivich2025systematic})—for example, predicting lower mastery even after a student answers correctly—resulting in sequences that contradict the expected gradual progression of learning \citep{hooshyar2025problems, yeung2018addressing}. \citet{wang2023wrong} investigated this behavior using finite-state automata and showed that such prediction volatility stems from structural properties of the model itself. Finally, DKT suffers from the well-known "black-box" problem that is its decision-making processes are opaque, making it difficult for educators and stakeholders to interpret or trust its outputs \citep{krivich2025systematic}. This lack of explainability is particularly concerning in educational settings, where opaque or biased systems may reinforce inequities and fail to support diverse learners effectively \citep{baker2022algorithmic}. To address these concerns, recent research has increasingly explored explainable AI approaches aimed at improving model transparency and trustworthiness \citep{arrieta2020explainable}. Post-hoc explanation techniques, such as SHAP and LIME, can provide insights into model predictions and offer some indication of the factors influencing AI decisions \citep{saarela2021explainable}. However, several studies suggest that these explanations are often incomplete or potentially misleading, limiting their effectiveness for achieving true interpretability and trust \citep{hooshyar2024problems, lakkaraju2020fool}.

\subsection{Responsible AI for education through hybrid neural-symbolic computing}
\label{subsec:responsible_ai}

As education is recognised as a \textit{high-risk} domain for AI deployment in the EU, issues of trust and interpretability extend beyond technical concerns and become ethical requirements \citep{europeanunion2024artificial}. Addressing these challenges calls for moving away from opaque, purely data-driven models toward hybrid approaches that integrate domain expertise, involve educators and stakeholders in their development, improve transparency, and align system behaviour with pedagogical and ethical principles. Responsible AI provides a foundational lens for this shift, emphasizing principles such as fairness, transparency, accountability, privacy, and human agency \citep{eitel2020beyond, jakesch2022different, maree2020towards, viberg2026protecting, werder2022establishing}. In this study, we subscribe to the definition of responsible AI proposed by \citet{goellner2024responsible}, which conceptualises responsible AI as "a \textbf{human-centred} approach aimed at fostering user \textbf{trust} through \textbf{ethical} and reliable decision-making, \textbf{explainable} outcomes, and \textbf{privacy}-preserving implementation". From a design perspective, this implies prioritising human-centred values, ethical and reliable decision processes, and explainable AI methods. Embedding these principles within model development can naturally strengthen user trust and promote privacy-preserving practices \citep{hooshyar2025towardsb, hooshyar2025towards}.

A promising direction for addressing the limitations of purely data-driven models is the development of hybrid human–AI systems that integrate training data with structured domain knowledge \citep{besold2021neural, hooshyar2025towards}. Within this line of work, neural-symbolic AI (NSAI)\footnote{While the term ``neural-symbolic'' has traditionally been used in the research community, ``neurosymbolic'' is also used interchangeably in academic literature and the media.}—often described as the ``\textit{third wave of AI}''—combines symbolic reasoning with data-driven learning, uniting the interpretability of expert rules with the predictive strength of neural networks \citep{garcez2023neurosymbolic, hooshyar2021neural}. In educational applications, NSAI supports a more collaborative and human-centred approach to model development. Domain experts, including educators and researchers, can incorporate their knowledge into the modelling process through direct injection of their symbolic knowledge \citep{hooshyar2024temporal, tato2022infusing}. Integrating such knowledge can also help reduce the effects of biased, incomplete, or noisy datasets by supplementing data-driven learning with expert-informed guidance \citep{hooshyar2024augmenting}. Furthermore, NSAI frameworks allow the inclusion of structured educational concepts—such as causal dependencies, learning theories, the role of specific variables, or their interactions—thereby improving the pedagogical grounding of AI-driven predictions \citep{shakya2021student}. Another key advantage is improved transparency, as neural-symbolic models often link predictions to explicit reasoning structures, making their decision processes more interpretable \citep{besold2021neural, hooshyar2024problems}. These characteristics also support key principles of responsible AI: strengthening user trust through transparent and pedagogically meaningful decisions \citep{nazaretsky2022teachers}; promoting ethical decision-making by basing predictions in educational knowledge rather than opaque statistical patterns \citep{holmes2022ethics}; and enabling privacy-preserving use by limiting dependence on sensitive student data—not only because explicit knowledge injection allows accurate modelling with less raw data, but also because rules related to ethics and data protection can be embedded directly into the model, ensuring that personal information becomes less identifiable and that system behaviour respects privacy by design \citep{porayska2023ethics}. Finally, by actively involving stakeholders in the development process, NSAI reinforces a human-centred approach to AI in education. This human-in-the-loop approach makes NSAI particularly appropriate for educational contexts, enhancing transparency, building trust, and promoting ethical responsibility \citep{hooshyar2025towards}.

\subsection{Objective and research questions}
\label{subsec:objectives}

Recent studies have begun to explore how neural-symbolic AI can enhance education by embedding domain knowledge into neural architectures. Examples include hybrid frameworks that combine symbolic representations of learner behaviours with deep neural networks for predicting learner strategies \citep{venugopal2021neuro}, Bayesian networks integrated with deep learning for learner modelling under data sparsity \citep{tato2022infusing}, and NSAI approaches that inject and extract educational knowledge into and from deep neural networks to model learners' performance \citep{hooshyar2024augmenting}. These approaches demonstrate the potential of NSAI to improve interpretability, generalizability, and fairness in educational AI, while remaining aligned with established theories of learning. Despite this promise, there is a lack of research on applying NSAI to sequential learner modelling approaches such as DKT, which remains a dominant method for modelling learner knowledge over time. Therefore, this study aims to develop a novel, responsible deep knowledge tracing approach (called \textbf{\textit{Responsible-DKT}}) that integrates symbolic practitioner knowledge into sequential neural architectures (using recurrent neural networks as a proof-of-concept example, while the approach is applicable to other architectures such as Transformers) for interpretable and trustworthy learner modelling. To this end, we employ the concept of Lifted Relational Neural Networks \citep{sourek2018lifted}, which enables the creation of differentiable logic programs for combining symbolic rules with deep learning. The study is structured around the following research questions:

\begin{itemize}
    \item \textbf{RQ1}: How does Responsible-DKT compare with conventional DKT in terms of predictive accuracy?
    \item \textbf{RQ2}: To what extent does symbolic knowledge injection in Responsible-DKT improve the sequential stability of predictions over time?
    \item \textbf{RQ3}: How does Responsible-DKT provide interpretable explanations of student knowledge predictions through its neural-symbolic computation structure?
\end{itemize}

The remainder of the paper presents related work (Section~\ref{sec:related_work}), the proposed method (Section~\ref{sec:method}), results and analysis (Section~\ref{sec:results}), and discussion and conclusions (Section~\ref{sec:discussion}).

\section{Related Work}
\label{sec:related_work}

\subsection{Deep knowledge tracing}
\label{subsec:dkt}

In digital learning environments, DKT has become a central approach for modelling learners' evolving knowledge. Introduced by \citet{piech2015deep}, DKT applied recurrent neural networks to analyse sequential learner interactions, capture temporal dependencies, and predict skill mastery over time—a breakthrough that inspired extensive follow-up research. Since then, numerous studies have extended DKT with novel architectures and training strategies to improve predictive accuracy \citep{krivich2025systematic}.

A central strength of deep knowledge tracing research is that it has produced multiple modelling traditions for representing learner knowledge over time in a structured and computationally explicit way. Recent reviews identify six major lines of development and synthesize their architectures, benchmark datasets, application areas, and open challenges \citep{krivich2025systematic, abdelrahman2023knowledge}. The first category is \textit{sequence modelling}, beginning with DKT by \citet{piech2015deep}, which used recurrent neural networks and long short-term memories to predict the probability of a correct response at each time step. Extensions include Extended-DKT \citep{xiong2016going}, which incorporated auxiliary student and exercise features; DKT+ \citep{yeung2018addressing}, which added regularization terms to improve reconstruction and reduce prediction inconsistencies; and DKT-DSC \citep{minn2018deep}, which dynamically clustered students based on performance using K-means. The second category is \textit{memory-augmented} models, which enhance DKT by incorporating external memory structures inspired by memory-augmented neural networks \citep{graves2014neural}. Unlike the hidden state in DKT, these models use a key–value memory to explicitly represent knowledge states: the key matrix encodes knowledge components (KCs), while the value matrix tracks a student's mastery level. A notable example is DKVMN \citep{zhang2017dynamic}, which introduced a dynamic value matrix to capture the evolving mastery of each KC, while keeping the key matrix static. Later, SKVMN \citep{abdelrahman2019knowledge} addressed limitations of DKT and DKVMN by introducing Hop-LSTM, a sequential modelling component that better captures dependencies among questions and updates student knowledge states based on responses to relevant KCs. The third category is \textit{attentive} models, which draw on the Transformer architecture \citep{vaswani2017attention} to assign explicit, learnable importance weights to past interactions. The first such model, SAKT \citep{pandey2019self}, used multi-head scaled dot-product attention to weigh past questions when predicting future responses. AKT \citep{ghosh2020context} extended this idea with monotonic attention, introducing an exponential decay to model forgetting, along with Rasch-based embeddings to capture deviations between questions and their associated skills. SAINT \citep{choi2020towards} applied a full Transformer encoder–decoder structure by separating question and response sequences, later enhanced with time-related features in SAINT+.

The fourth category of deep knowledge tracing is \textit{graph-based} models, which leverage graph neural networks (GNNs) to capture relational structures such as similarity and dependency among KCs, as well as question–KC correspondences. GKT \citep{nakagawa2019graph} reformulated knowledge tracing as a time-series node classification task, where KCs are nodes and their dependencies are edges, using message-passing GNNs with both statistics-based and learned graph construction methods. GIKT \citep{yang2020gikt} extended this by modelling many-to-many relations between questions and KCs, aggregating embeddings through GNNs before feeding them into an recurrent neural networks for prediction. SKT \citep{tong2020structure} further captured multiple KC relations, such as similarity and prerequisite structures, combining temporal and spatial graph embeddings to improve predictions. The fifth category is \textit{text-aware} models, which incorporate the textual content of questions to improve representation learning and prediction. EERNN \citep{su2018exercise} used a bi-directional long short-term memory to extract embeddings from question text and combined them with interaction histories through another long short-term memory. \citet{yin2019quesnet} extended this approach with a masked language modelling pre-training step to enhance question representations. EKT \citep{liu2019ekt} further advanced text-aware KT by representing a student's knowledge state as a matrix over multiple KCs, using a memory network to quantify how each question influences mastery. The sixth category is \textit{forgetting-aware} models, which explicitly account for the decline of knowledge mastery over time, inspired by learning psychology and Ebbinghaus's forgetting curve \citep{ebbinghaus2013memory}. DKT+Forgetting \citep{nagatani2019augmenting} extended DKT by incorporating attributes such as the number of previous attempts, the time elapsed since the learner last interacted with the same knowledge component, and the time since the learner's most recent interaction with any task, allowing the model to better represent both learning progress and forgetting over time. HawkesKT \citep{wang2021understanding} leveraged a Hawkes point process \citep{hawkes1971spectra} to model temporal cross-effects, recognizing that forgetting rates vary across KCs. DGMN \citep{abdelrahman2022deep} combined GNNs with dynamic memory to model KC relationships and forgetting features, integrating them via an attention–gating mechanism.

In \citeauthor{krivich2025systematic}'s (\citeyear{krivich2025systematic}) work—the first systematic review of the area—the authors examined deep knowledge tracing research from the perspective of responsible AI. Their findings reveal that less than half of the reviewed works address data quality issues during model development, only a small minority evaluate the sequential stability of predictions (i.e., the consistency of predictions over time), and the majority of studies do not provide interpretability for their predictions and decision-making processes. As is apparent, several challenges remain unresolved, which may lead to limited alignment with pedagogical principles and reduce the reliability and broader adoption of these models in educational settings.

\subsection{Neural-symbolic AI for education}
\label{subsec:nsai}

Neural-symbolic computing is an approach that combines the reasoning capabilities of symbolic AI with the learning power of neural networks. This integration allows models to benefit from both structured knowledge and data-driven learning, bringing together the transparency of symbolic systems and the predictive strength of deep learning \citep{besold2021neural, garcez2023neurosymbolic, sourek2018lifted}. \citet{kautz2022third} described six main strategies for combining these paradigms, including using neural networks to process symbolic inputs \citep{pennington2014glove}, integrating neural components within symbolic systems such as AlphaGo \citep{silver2016mastering}, transforming raw data into symbolic representations for reasoning \citep{mao2019neuro}, guiding neural learning with symbolic rules \citep{lample2019deep}, embedding rules directly into neural architectures \citep{serafini2017learning}, and incorporating symbolic reasoning modules inside neural models \citep{kahneman2011thinking}. These approaches illustrate the flexibility of neural-symbolic AI and its potential to address complex challenges across different domains.

In educational contexts, hybrid neural-symbolic approaches have the potential to align algorithmic predictions with pedagogical theories, enhance generalization, improve interpretability, and mitigate bias—key requirements for responsible and trustworthy AI in education \citep{hooshyar2025towards, hooshyar2021neural, tato2022infusing, venugopal2021neuro}. In recent years, researchers have begun applying NSAI in educational settings. \citet{shakya2021student} integrated Markov Logic Networks with long short-term memories to embed domain knowledge for predicting student strategies, demonstrating improved efficiency and reduced overfitting on KDD EDM challenge datasets. \citet{tato2022infusing} proposed a hybrid model that combined Bayesian networks with deep learning to address data inconsistency, which improved predictive accuracy on sparse or imbalanced learner data. Moreover, a series of studies by Hooshyar and colleagues have explored the potential of NSAI to enable responsible and trustworthy AI applications in education \citep{hooshyar2024temporal, hooshyar2024augmenting, hooshyar2025towardsa, hooshyar2024problems}. Their work has shown that embedding educational causal relationships into neural networks improves generalizability and enables the extraction of human-readable rules \citep{hooshyar2024augmenting}, that NSAI-based models provide more faithful and pedagogically aligned explanations than post-hoc methods such as SHAP and LIME \citep{hooshyar2024problems}, and that incorporating educational principles directly into loss functions of autoencoders can guide unsupervised models to penalize behaviors inconsistent with pedagogical rules while improving generalization \citep{hooshyar2024temporal}. More recently, \citet{hooshyar2025towardsa} compared (sub)symbolic and neural-symbolic approaches in educational data mining, using self-regulated learning data to predict students' mathematics performance and uncover influential learning factors. Their findings showed that NSAI offered a stronger balance of generalizability and interpretability while addressing data imbalance and providing actionable insights. 

\textit{In summary, existing works highlight the potential of NSAI to move beyond accuracy-driven approaches toward models that are interpretable, theory-driven, and trustworthy, thereby contributing to the development of responsible AI for education}. However, despite these advances, the application of neural-symbolic AI approaches to sequential models for student knowledge tracing remains largely unexplored. In particular, little work has investigated how symbolic educational knowledge can be integrated into temporal interaction modelling frameworks such as deep knowledge tracing.

\subsection{Relational neural-symbolic learning}
\label{subsec:relational_nsai}

While early neural-symbolic AI has successfully integrated symbolic knowledge with deep machine learning, much of this work operates at the so-called \textit{propositional} level. Propositional neural-symbolic models, restricted to describing relationship between specific input features, typically rely on basic neural architectures, such as Multilayer Perceptrons (MLPs), inherently requiring fixed-size inputs. Consequently, they struggle to natively handle complex, structured, or dynamically sized data. To address these limitations, this work leverages the Lifted Relational Neural Networks (LRNN) paradigm \citep{sourek2018lifted}, implemented via the ``PyNeuraLogic'' framework\footnote{\url{https://github.com/LukasZahradnik/PyNeuraLogic/}}. PyNeuraLogic, which naturally supports relational and temporal reasoning required for modelling sequential learner interactions, elevates neural-symbolic learning to the relational level, utilizing differentiable first-order logic programming. This paradigm is inherently designed to process structured relational representations, such as knowledge graphs and relational databases. Because temporal sequences are fundamentally a special, linear case of relational graphs, PyNeuraLogic can subsume the capabilities of structured neural architectures, like Graph or Recurrent Neural Networks (RNNs), while simultaneously enforcing explicit symbolic rules.

Within the PyNeuraLogic framework, information is not rigidly formatted into numerical matrices, but is instead expressed naturally using logical \textit{predicates} that define \textit{relationships} between entities. When these abstract predicates are populated with actual observations from a dataset—such as a specific student answering a specific quiz item correctly at a given timestep—they become \textit{ground facts}, forming a ``contextual knowledge base''. To process this knowledge base, the model's architecture is declared as a \textit{template}, which is a compact set of parameterized, weighted logical rules. Because these rules are ``lifted''—containing abstract relational variables rather than fixed entities—a single learning template can generalize across data structures and sequences of arbitrary length and complexity.

To learn from the data, the PyNeuraLogic framework dynamically translates this abstract template into a trainable neural network through a process called \textit{grounding}. Grounding algorithmically matches the lifted rules against the available ground facts to unroll a directed acyclic computation graph uniquely tailored to the specific structure of the input data. The model is then optimized to evaluate a specific \textit{query}—the target logical statement it seeks to predict, such as a student's future performance—using standard backpropagation through the fully differentiable grounded graph. This relational approach is then well-suited for temporal student interaction modeling. By treating the learning sequence as a relational structure, the LRNN paradigm allows us to write logic templates that simultaneously compute recurrent neural state transitions and enforce dynamic pedagogical constraints across time steps. Building upon this declarative foundation, the following section introduces our novel sequential architecture that explicitly encodes these temporal educational interactions into a unified, differentiable logic program.

\section{Method}
\label{sec:method}

\begin{figure}[t]
\centering
\input{figure/nesy_dkt_workflow}
\caption{Architecture of the proposed Responsible-DKT approach.}
\label{fig:fig1}
\end{figure}
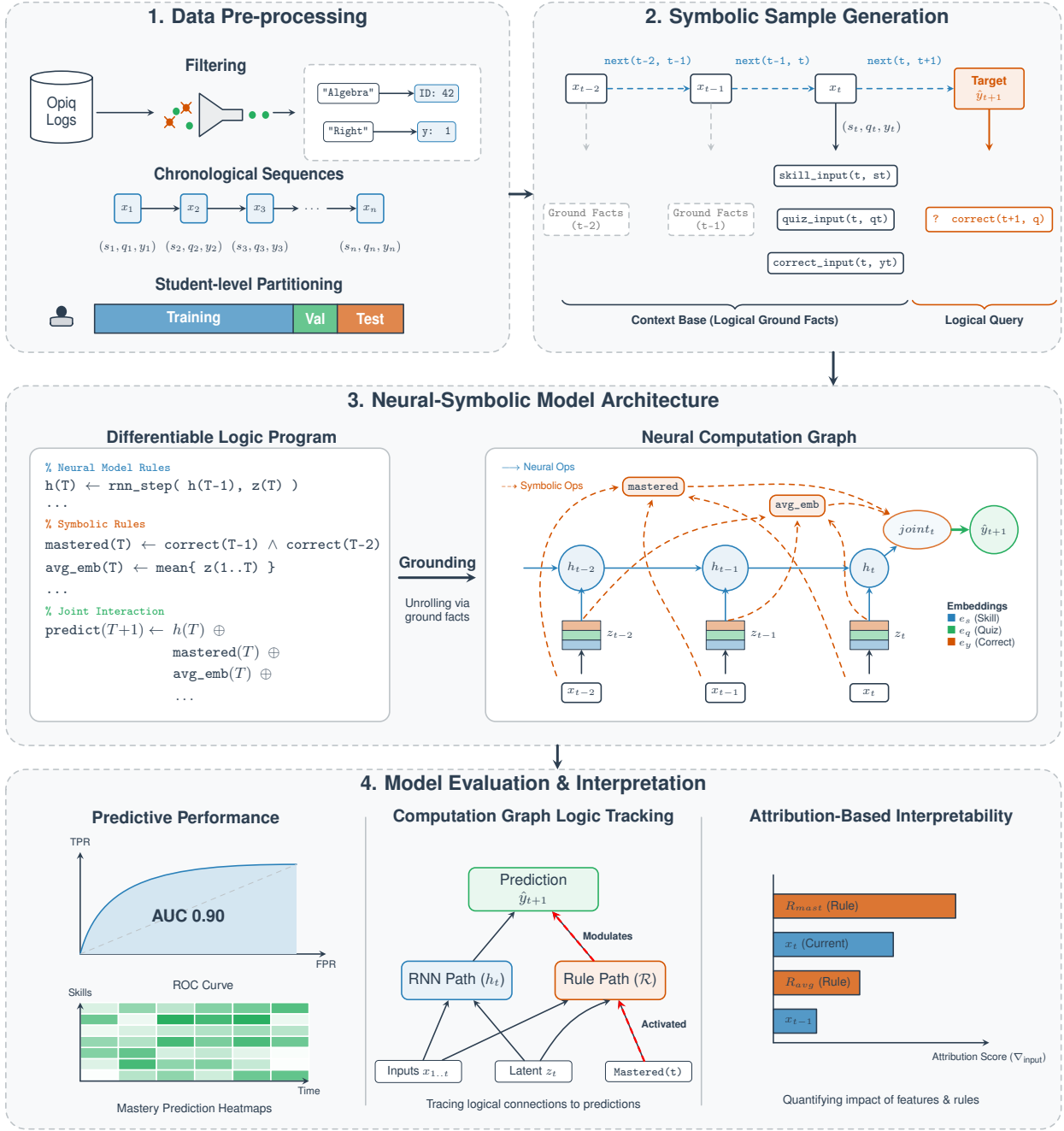

Building upon the declarative relational framework established in Section~\ref{subsec:relational_nsai}, our approach, \textbf{Responsible-DKT}, is a neural-symbolic variant of DKT that injects symbolic educational knowledge into sequential neural architectures. By formulating the model as a differentiable logic program, it also allows us to extract knowledge from the trained network to provide interpretable insights into predictions. In this work, we focus on incorporating simple pedagogical rules (e.g., mastery and non-mastery heuristics; \citealp{koedinger2023astonishing}) as an illustrative example, while the framework is designed to support the integration of more complex forms of educational knowledge (see the Discussion section). As shown in Fig.~\ref{fig:fig1}, the approach includes four main components: data pre-processing, sample generation, neural-symbolic modelling, and model evaluation and interpretation. In \textit{data pre-processing} (Fig. \ref{fig:fig1}-1), raw interaction logs are pre-processed and encoded to construct ordered per-student interaction sequences (skill, quiz, correctness). In \textit{symbolic sample generation} (Fig. \ref{fig:fig1}-2), a sequence-to-sequence learning formulation is adopted in which symbolic inputs are incrementally constructed and next-step predictions are defined within each student sequence. The proposed \textit{neural-symbolic model} (Fig. \ref{fig:fig1}-3), implemented in PyNeuraLogic (Sec.~\ref{subsec:relational_nsai}), integrates embedding and recurrent components with explicit symbolic rules (e.g., \textit{mastery} and \textit{non-mastery} constraints based on consecutive responses). These components are used here as a proof-of-concept, although more complex architectures such as Transformers and richer symbolic rules could also be employed, enabling sequential neural-symbolic reasoning over student learning trajectories. \textit{Model evaluation and interpretation} (Fig. \ref{fig:fig1}-4) involve both performance analysis and interpretability. Performance is assessed using standard predictive metrics (e.g., AUC), complemented by early–middle–late error comparison across different stages of the student sequence and temporal coherence analyses, including prediction volatility and inconsistency. Qualitative analyses, including skill-level mastery heatmaps, further support interpretation of predictive behaviour over time. Our proposed approach is compared, using the aforementioned metrics, with a baseline version of the neural-symbolic model without knowledge injection (\textbf{BaseNS-DKT}, Base Neural-Symbolic Knowledge Tracing) and with its equivalent fully data-driven PyTorch baseline (\textbf{Classic-DKT}). Interpretability analyses comprise both local and global explanations for the Responsible-DKT method, as it provides inherently interpretable decision-making logic through its symbolic components. In contrast, the baseline models are primarily data-driven and require post-hoc explanation methods, making direct comparison less meaningful. Algorithm~\ref{alg:framework} summarizes the methodological framework, and Table~\ref{tab:notation} defines the notations used.

\begin{table}[t]
\centering
\caption{Notation.}
\label{tab:notation}
\begin{tabular}{ll}
\toprule
\textbf{Symbol} & \textbf{Meaning} \\
\midrule
$\mathbb{D}$ & Student interaction dataset \\
$U \in \{u^1, u^2, \dots, u^{|U|}\}$ & Set of students \\
$Q \in \{q^1, q^2, \dots, q^{|Q|}\}$ & Set of quizzes \\
$S \in \{s^1, s^2, \dots, s^{|S|}\}$ & Set of skills \\
$t$ & Discrete timestep within a student's interaction sequence \\
$x_t = (s_t, q_t, y_t)$ & Input tuple with skill, quiz, and correctness at timestep $t$ \\
$e_q, e_s, e_y$ & Embeddings of quiz, skill, and correctness inputs \\
$h_t$ & Hidden state of the recurrent neural network at timestep $t$ \\
$\hat{y}_{(t+1,q)}$ & Predicted probability of correctness for quiz $q$ at time $t+1$ \\
$y_{(t+1,q)} \in \{0,1\}$ & Ground truth correctness label (0/1) \\
$\mathcal{R}$ & Set of symbolic rules (e.g., mastery/not-mastery constraints) \\
BCE & Binary Cross-Entropy loss \\
\bottomrule
\end{tabular}
\end{table}

\begin{algorithm}[t]
\caption{Neural-symbolic knowledge tracing framework and evaluation pipeline.}
\label{alg:framework}
\begin{algorithmic}[1]
\REQUIRE Interaction dataset $\mathbb{D} = \{(u, t, x_t)\}$, where $x_t = (s_t, q_t, y_t)$
\ENSURE Trained models, evaluation metrics, and interpretability artifacts

\STATE \textbf{Pre-processing}
\STATE Remove interactions with missing identifiers
\STATE Encode skills ($s \in S$), quizzes ($q \in Q$), and students ($u \in U$) as categorical indices
\STATE Convert scores to binary correctness labels $y_t \in \{0,1\}$ using a predefined threshold
\STATE Sort interactions by student and time; assign discrete timestep indices $t$
\STATE Construct ordered per-student interaction sequences $x_t = (s_t, q_t, y_t)$
\STATE Split students into train/validation/test sets at the student level

\STATE \textbf{Sample generation}
\FOR{each student sequence}
    \STATE Generate incremental training samples using the interaction history up to timestep $t$
    \STATE Build a sequence-to-sequence formulation:
    \STATE \quad Inputs at timestep $t$ consist of symbolic facts: skill\_input$(t, s_t)$, quiz\_input$(t, q_t)$, correct\_input$(t, \{\text{right}, \text{wrong}\})$
    \STATE \quad Queries predict next-step correctness at $t+1$
    \STATE \quad Encode temporal transitions between consecutive interactions
\ENDFOR

\STATE \textbf{Modelling}
\STATE \textit{Knowledge-augmented neural-symbolic DKT (Responsible-DKT)}
\STATE \quad Embedding layer for quizzes, skills, and correctness inputs
\STATE \quad Input embeddings are combined into a unified timestep representation
\STATE \quad Recurrent core (RNN/LSTM) producing hidden states $h_t$
\STATE \quad Output layer predicting $\hat{y}_{t+1}$ using the recurrent state and target quiz/skill context
\STATE \quad Learnable-weight symbolic rules $\mathcal{R}$ injected into the model to influence predictions
\STATE \textit{Base neural-symbolic model without knowledge injection (BaseNS-DKT)}
\STATE \quad Identical neural-symbolic architecture with the symbolic rule set $\mathcal{R}$ disabled
\STATE \textit{PyTorch baseline (Classic-DKT)}
\STATE \quad Fully data-driven RNN-based DKT implemented in PyTorch

\STATE \textbf{Training}
\STATE Optimize binary cross-entropy $BCE(\hat{y}_{t+1}, y_{t+1})$ using Adam (learning rate $10^{-4}$)
\STATE Train for up to 300 epochs with early stopping based on validation loss
\STATE Use the best validation model for final testing

\STATE \textbf{Evaluation}
\STATE Compute predictive metrics: AUC, Accuracy, Precision, Recall, and F1 score
\STATE Early–middle–late sequence error analysis and temporal coherence analysis
\STATE Qualitatively examine the temporal stability of model predictions across student sequences

\STATE \textbf{Interpretability (Responsible-DKT)}
\STATE Inspect grounded neural-symbolic computation graphs to trace the main predictive pathways
\STATE Local explanations: Gradient–value attribution of input facts aggregated per timestep
\STATE Global explanations:
\STATE \quad Rank skill and quiz embeddings by aggregated gradient importance
\STATE \quad Analyze dataset-level influence of symbolic rules $\mathcal{R}$ using average rule activation $|val|$ and gradient sensitivity $|grad|$
\STATE \quad Visualize relative skill importance over interaction timesteps using gradient-based heatmaps
\end{algorithmic}
\end{algorithm}

\subsection{Dataset and preprocessing}
\label{subsec:dataset}

We used a dataset consisting of 6th-grade mathematics interaction logs from Estonia, collected during regular classroom use in September 2021. The dataset provides time-ordered interaction sequences that support the development of sequential models and the evaluation of predictive performance and sequential stability. The data originate from Opiq\footnote{\url{https://www.opiq.ee/Catalog}}, a nationally adopted K–12 digital learning platform widely used in Estonia for mathematics and other school subjects. Opiq supports curriculum-aligned practice and assessment and logs fine-grained learner interaction data during authentic educational use. The dataset contains 21,471 student–item interactions, each originally described by more than 20 raw attributes. The records capture individual student problem-solving attempts and include identifiers and performance information such as student\_id, quiz\_id, skill\_id, and the response outcome. Table~\ref{tab:dataset_stats} summarizes the main characteristics of the dataset. Student performance scores were binarized using the first quartile (score = 37) of the score distribution as the threshold: scores below the threshold were labelled incorrect (0), and those at or above it was labelled correct (1). This discretization allows the identification of lower-performing learners falling into the bottom 25\% of observed performance. Interactions were sorted chronologically for each student and indexed with a per-student timestep. Problem and skill identifiers were encoded as categorical indices to support sequential modeling. Learners who had fewer than two recorded interactions were removed from the dataset, as at least two interactions are required to support sequential modelling and next-step prediction. The resulting data consist of time-ordered learner interaction sequences, where each interaction is represented as a triplet $(s_t, q_t, y_t)$ corresponding to skill, quiz, and correctness, capturing the learner's evolving performance over time. The data were partitioned by student into training, validation, and test sets with an approximate 7:1:2 ratio, ensuring that each learner appears in only one subset. This student-level partitioning prevents information leakage and enables evaluation on previously unseen learners. Models (Responsible-DKT, BaseNS-DKT, and Classic-DKT) were trained and evaluated on a next-step prediction task, in which correctness at the current interaction $t$ is predicted using the learner's interaction history up to $t-1$. This formulation follows standard DKT practice and ensures fair comparison across methods using identical interaction trajectories.

\begin{table}[t]
\centering
\caption{Descriptive statistics of the dataset.}
\label{tab:dataset_stats}
\begin{tabular}{lrrrr}
\toprule
\textbf{Statistic} & \textbf{Original} & \textbf{Train} & \textbf{Val} & \textbf{Test} \\
\midrule
\# of records (interactions) & 21471.0 & 14057.0 & 2969.0 & 4428.0 \\
\# of students & 167.0 & 105.0 & 15.0 & 30.0 \\
\# of quizzes & 1058.0 & 939.0 & 565.0 & 647.0 \\
\# of skills & 13.0 & 13.0 & 12.0 & 13.0 \\
Avg. interactions per student & 128.57 & 133.88 & 197.93 & 147.6 \\
Avg. interactions per quiz & 20.29 & 14.97 & 5.25 & 6.84 \\
Avg. interactions per skill & 1651.62 & 1081.31 & 247.42 & 340.62 \\
Correct interactions (y=1) & 16118.0 & 10565.0 & 2231.0 & 3305.0 \\
Incorrect interactions (y=0) & 5353.0 & 3492.0 & 738.0 & 1123 \\
\bottomrule
\end{tabular}
\end{table}

\begin{figure}[t]
\centering
\includegraphics[width= 400pt]{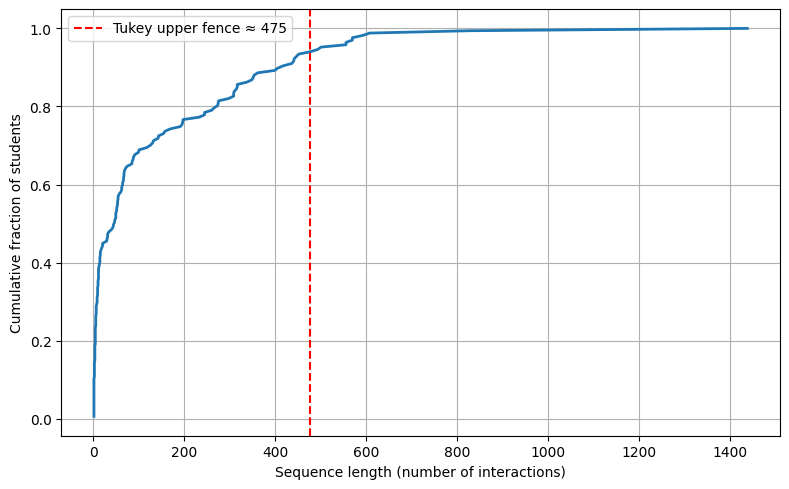}
\caption{Cumulative distribution function of student interaction sequence lengths.}
\label{fig:fig2}
\end{figure}

To control for extreme variation in student interaction histories, we analyzed the distribution of sequence lengths after preprocessing. Sequence-length statistics are summarized in Table~\ref{tab:sequence_stats}. As shown in the table, the distribution is highly right-skewed, with a small number of students exhibiting very long interaction histories. Using Tukey's rule for outlier detection \citep{tukey1977exploratory}, the upper fence was computed as $Q3+1.5(Q3-Q1)$, which corresponds to approximately 475 interactions given $Q1=5$ and $Q3=193$. This threshold identifies 11 students (6.59\% of the cohort) as having unusually long sequences. Rather than excluding these students, we retained all learners and limited sequence length during training by truncating each student’s history to a maximum of 475 interactions, thereby preventing extremely long sequences from dominating the training process. We then evaluated multiple maximum sequence-length conditions to study model behavior across different learning regimes. Specifically, we considered 10 interactions to represent cold-start scenarios with minimal prior information, as this value is close to the lower quartile ($Q1 = 5$) while allowing sufficient history for next-step prediction after the first interactions. We selected 50 interactions to capture early learning trajectories, as this value closely reflects the median sequence length (46) and therefore represents a typical student history. Finally, 100 and 475 interactions were used to reflect progressively longer histories: 100 interactions exceed the median and approach the mean sequence length ($\approx$129), while 475 corresponds to the Tukey upper fence, marking the upper bound of the non-outlier distribution. As illustrated by the cumulative distribution of sequence lengths in Fig.~\ref{fig:fig2}, the vast majority of learners have interaction histories well below the Tukey upper fence, and fewer than 7\% of students are excluded by this criterion. The chosen sequence-length limits therefore preserve representative learning trajectories while systematically covering sparse, typical, and extended learning histories.

\begin{table}[t]
\centering
\caption{Summary statistics of student interaction sequence lengths.}
\label{tab:sequence_stats}
\begin{tabular}{lr}
\toprule
\textbf{Sequence length statistic} & \textbf{Value} \\
\midrule
Min & 1 \\
Q1 & 5.0 \\
Median & 46.0 \\
Mean & 128.57 \\
Q3 & 193.0 \\
Max & 1437 \\
\bottomrule
\end{tabular}
\end{table}

\subsection{Sample generation}
\label{subsec:sample_gen}

From the pre-processed dataset, training examples were generated at the level of individual students. For each student $u$, interactions were first sorted chronologically and indexed by discrete timesteps $t$. Each interaction was represented as a tuple:
\begin{equation}
x_t = (s_t, q_t, y_t)
\label{eq:interaction_tuple}
\end{equation}
where $s_t$ denotes the exercised skill, $q_t$ the corresponding quiz item, and $y_t \in \{0, 1\}$ the observed correctness. Subsequently, instead of translating the interaction logs into standard one-hot encoded tensors, we encode the pre-processed student interactions into symbolic ground facts (Section~\ref{subsec:relational_nsai}). Each timestep in a student's sequence is translated into a set of logical predicates mapping to the respective skill, quiz, and correctness. Particularly, at each timestep $t$, each interaction was encoded using symbolic predicates in PyNeuraLogic (Sec.~\ref{subsec:relational_nsai}) as:
\begin{equation}
\text{skill\_input}(t, s_t), \text{quiz\_input}(t, q_t), \text{correct\_input}(t, \{\text{right}, \text{wrong}\})
\label{eq:symbolic_predicates}
\end{equation}

Temporal dependencies were explicitly represented using grounded transition facts:
\begin{equation}
\text{next}(t, t+1)
\label{eq:temporal_transition}
\end{equation}

The prediction task was formulated as next-step knowledge tracing. Given interaction history up to timestep $t$, the model predicts correctness at timestep $t+1$. Depending on the output context, predictions were indexed by either skills or quizzes:
\begin{equation}
\hat{y}(t+1, s) = \text{correct}(t+1, s)
\label{eq:skill_prediction}
\end{equation}
\begin{equation}
\hat{y}(t+1, q) = \text{correct}(t+1, q)
\label{eq:quiz_prediction}
\end{equation}

Target predictions were formulated as logical queries (Sec.~\ref{subsec:relational_nsai}) generated for each next-step transition within the interaction sequence. For each student, a single multi-query sequence-to-sequence sample was constructed, containing one query for each valid next-step prediction $t+1$. Each sample consists of a symbolic encoding of the full interaction history and a set of queries corresponding to all valid next-step predictions.

\subsection{Model architecture}
\label{subsec:architecture}

\subsubsection{Knowledge-augmented neural-symbolic DKT (Responsible-DKT)}
\label{subsubsec:responsible_dkt}

The Responsible-DKT model is implemented as a single neural-symbolic template (Sec.~\ref{subsec:relational_nsai}) that jointly integrates symbolic representations, neural sequence modelling, and pedagogical rules $\mathcal{R}$ within the PyNeuraLogic framework. During training, this template is grounded (Sec.~\ref{subsec:relational_nsai}) against the students' specific interaction facts to unroll the computation graph.

At each timestep $t$, a student interaction $x_t = (s_t, q_t, y_t)$ is encoded as symbolic facts, as shown in equation~\ref{eq:symbolic_predicates}. Temporal order is explicitly grounded using transition facts $\text{next}(t, t+1)$, enabling reasoning across consecutive interactions. Each quiz $q \in Q$, skill $s \in S$, and correctness outcome is associated with a learnable embedding:
\begin{equation}
e_q, e_s, e_y \in \mathbb{R}^d
\label{eq:embeddings}
\end{equation}

For each timestep $t$, embeddings corresponding to \textit{quiz\_input}, \textit{skill\_input}, and \textit{correct\_input} are combined into a unified latent representation:
\begin{equation}
z_t = \text{combined\_embed}(t)
\label{eq:combined_embed}
\end{equation}

Correctness embeddings are modelled separately to prevent label leakage and to preserve interpretability of outcome effects. The combined embedding sequence $\{z_t\}$ is processed by a stacked recurrent core (RNN). Hidden states are updated via recurrent rule modules,
\begin{equation}
h_t = f(z_{t-1}, h_{t-1})
\label{eq:hidden_state}
\end{equation}

The final neural representation used for prediction at timestep $t$ is obtained through an explicit temporal shift:
\begin{equation}
\text{final\_nn}(t) \leftarrow \text{rnn\_out}(t-1) \wedge \text{next}(t-1, t)
\label{eq:temporal_shift}
\end{equation}

This design ensures that predictions at $t$ depend only on past interactions. Depending on the output context, the model predicts next-step correctness indexed by skill or quiz:
\begin{equation}
\hat{y}_t(s) = \text{correct}(t, s), \quad \hat{y}_t(q) = \text{correct}(t, q)
\label{eq:prediction_output}
\end{equation}

The output combines the recurrent state $\text{final\_nn}(t)$ with the relevant target embedding ($e_s$ or $e_q$) and applies a sigmoid activation to produce a probability.

On top of the neural backbone, symbolic rules $\mathcal{R}$ encode pedagogically motivated constraints. A recent study highlights the importance of maintaining temporally consistent mastery estimates in sequential learner modelling, particularly in settings characterised by class imbalance and sparsely practiced skills, where purely data-driven models may exhibit unstable or counterintuitive prediction dynamics \citep{hooshyar2025problems}. To address these challenges, we introduce explicit symbolic $\text{mastered}$ and $\text{not\_mastered}$ states that modulate prediction confidence when simple evidence patterns occur in a learner's interaction history. Specifically, three incorrect responses on the same skill or quiz activate a $\text{not\_mastered}$ rule, providing a conservative signal that discourages premature overestimation of mastery. Conversely, two consecutive correct responses activate a $\text{mastered}$ rule, reflecting minimal positive evidence of learning without assuming full mastery from a single success. While prior work on mastery learning suggests higher thresholds (e.g., seven learning opportunities to master a typical knowledge component; e.g., \citep{koedinger2023astonishing}), we adopt lower thresholds to better suit the shorter and sparser interaction sequences in our dataset. These constraints help mitigate common DKT instabilities, where mastery probabilities may increase despite repeated incorrect answers or decrease following consistent correct responses \citep{krivich2025systematic, yeung2018addressing}. When triggered, these rules contribute additional evidence to the correctness predicate:
\begin{equation}
\text{correct}(t, x) \leftarrow \text{mastered}(x, t)
\label{eq:mastered_rule}
\end{equation}
\begin{equation}
\text{correct}(t, x) \leftarrow \text{not\_mastered}(x, t)
\label{eq:not_mastered_rule}
\end{equation}
where $x$ denotes either a skill or a quiz depending on the output context. Rule weights are learnable, allowing the strength of each symbolic influence to be adapted from data while preserving an interpretable rule structure. Alternatively, they may be fixed to enforce stronger pedagogical constraints when desired. Importantly, these symbolic components modulate rather than override the neural predictions. In addition, the model incorporates a symbolic historical aggregation pathway ($\text{avg\_embed}$) that summarizes past interaction evidence. For each skill or quiz $x$, previous timestep embeddings associated with that concept are aggregated:
\begin{equation}
\text{avg\_embed}(t, x) = \text{AVG}_{i<t} \{\text{combined\_embed}(i) \mid x_i = x\}
\label{eq:avg_embed}
\end{equation}

This aggregated representation captures accumulated interaction history and contributes additional context to the prediction:
\begin{equation}
\text{correct}(t, x) \leftarrow \text{avg\_embed}(x, t)
\label{eq:avg_embed_rule}
\end{equation}

All components—symbolic inputs, embeddings, recurrent dynamics, output prediction, and symbolic rules—are unified within a single differentiable template and optimized end-to-end using binary cross-entropy loss. Together, these symbolic rules incorporate simple but meaningful learning signals from interaction history, combining short-term mastery evidence with aggregated historical context. This integration helps stabilize prediction dynamics and supports more pedagogically consistent knowledge tracing while maintaining interpretability through $\mathcal{R}$.

\subsubsection{Base neural-symbolic model without knowledge injection (BaseNS-DKT)}
\label{subsubsec:basens_dkt}

To isolate the contribution of symbolic rule-based guidance, we introduce an ablation baseline in which the influence of the rule set $\mathcal{R}$ is removed while preserving the full neural-symbolic architecture. Specifically, the model retains the same symbolic input encoding, learnable embeddings ($e_q, e_s, e_y$), recurrent sequence modelling producing hidden states $h_t$, and sigmoid-based prediction of next-step correctness $\hat{y}_{t,(q)}$. In this variant, the symbolic rule module $\mathcal{R}$ is excluded from the template, so predictions are determined solely by the neural components. This controlled ablation allows performance differences to be attributed directly to the presence or absence of rule-based guidance, rather than to architectural changes.

\subsubsection{Classic PyTorch baseline (Classic-DKT)}
\label{subsubsec:classic_dkt}

While a wide variety of DKT variants have been proposed in the literature \citep{krivich2025systematic}, differing in architectural complexity and inductive biases, we implement a classic DKT baseline to serve as a transparent and well-controlled point of comparison. This choice ensures architectural and data-processing consistency with our neural-symbolic models, allowing differences in performance and behaviour to be attributed to the presence or absence of explicit symbolic structure rather than to confounding design choices. The Classic-DKT baseline was implemented in PyTorch using a purely neural recurrent architecture, without symbolic representations or rule-based constraints. The learning trajectory of each student was encoded as a time-ordered sequence of interaction triplets according to Equation~\ref{eq:interaction_tuple}. For a controlled comparison, the model was aligned with the Responsible-DKT backbone in terms of embedding dimension and recurrent type (see Sec~\ref{subsec:experimental_setting}).

At each timestep, the skill identifier, quiz identifier, and correctness outcome were mapped to learnable embeddings and combined through linear projections to form the input representation:
\begin{equation}
z_t = W_s e_{st} + W_q e_{qt} + W_a e_{yt}
\label{eq:classic_input}
\end{equation}
where $e_{st}$, $e_{qt}$, and $e_{yt}$ denote the embeddings for skill, quiz, and correctness, respectively. The resulting embedding sequence was processed by a recurrent neural network, producing hidden states that summarize the learner's interaction history. For each timestep $t$, the model predicts the probability of correctness for the next exercised item by combining the hidden state with the embedding of the target skill or quiz:
\begin{equation}
\hat{y}_{t+1} = \sigma(W h_t + e_{x_{t+1}})
\label{eq:classic_output}
\end{equation}
where $e_{x_{t+1}}$ denotes the embedding of the next target skill or quiz depending on the prediction context. Following the standard next-step DKT formulation, correctness at time $t+1$ is predicted using the interaction history up to timestep $t$. The model was optimized end-to-end using binary cross-entropy loss with the Adam optimizer and early stopping based on validation loss. Unlike the neural-symbolic variants, this Classic-DKT model relies entirely on learned embeddings and recurrent dynamics, without explicit symbolic constraints, or pedagogical rules. It therefore serves as a strong data-driven baseline for evaluating the benefits of explicit temporal reasoning and rule-based guidance introduced in Responsible-DKT.

\subsection{Model evaluation and interpretation}
\label{subsec:evaluation}

\subsubsection{Evaluation}
\label{subsubsec:evaluation_metrics}

Model performance was evaluated on a held-out test set using the standard next-interaction prediction task, where each instance represents a learner's activity at time $t+1$ with the true correctness label and the model's predicted probability of a correct response. Performance was assessed using common classification metrics, including AUC, accuracy, precision, recall, and F1-score, while additional analyses examined how model behaviour evolves across learning sequences \citep{hooshyar2025problems, krivich2025systematic}. Prediction errors were examined across different stages of the student interaction sequence by dividing each sequence into early, middle, and late segments. \textit{Stage-wise error} was computed as the percentage of misclassified instances in each segment using a fixed decision threshold of 0.5, aggregated across all students. Furthermore, \textit{volatility} was used as a smoothness-based measure, calculated as the mean absolute change in predicted probability between consecutive attempts on the same skill and aggregated across all student–skill transitions to evaluate the temporal coherence of predicted trajectories,
\begin{equation}
\frac{1}{N} \sum |P_{t} - P_{t-1}|
\label{eq:volatility}
\end{equation}
reflecting how stable or abrupt probability trajectories are. Also, \textit{inconsistency} rate was computed to quantify how often updates in predicted probabilities moved opposite to what would be expected given the correctness of the subsequent response. For instance, when predicted mastery increased after an incorrect answer—measured as the proportion of sign mismatches between the probability change and the expected update direction implied by the ground-truth label.
\begin{equation}
\Delta P_t = P_t - P_{t-1}
\label{eq:inconsistency}
\end{equation}

Finally, \textit{multi-skill mastery heatmaps} based on each model's predictions were used to analyse sequential prediction stability.

\subsubsection{Interpretability}
\label{subsubsec:interpretability}

Interpretability analyses were conducted only for the Responsible-DKT model and not for the Classic-DKT baseline. This reflects the neural-symbolic architecture's ability to expose explicit symbolic representations, grounded temporal structure, and interpretable computational graph. In contrast, Classic-DKT relies on dense neural states and implicit recurrence, providing limited access to intermediate reasoning processes. The neural-symbolic model therefore enables inspection of predictions as well as intermediate symbolic and neural components across time. Interpretability analyses comprise both \textit{local} and \textit{global} explanations. Local explanations attribute individual predictions to specific interaction timesteps by quantifying the contribution of symbolic input facts using gradient-based importance scores, which are mapped back to the original interaction sequence for step-wise interpretation. Global explanations aggregate importance scores across samples to identify influential skills, quizzes, and symbolic components, revealing overall model reliance patterns and their association with correctness outcomes. To illustrate model reasoning, representative prediction cases were first inspected through the grounded \textit{computation graph}. Local explanations were then generated for selected timesteps using gradient $\times$ input attribution, revealing how previous interactions supported or opposed the predicted outcome. At the global level, aggregated importance scores were used to identify influential skills and quizzes across the dataset, followed by \textit{rule-level analysis} examining the activation and gradient sensitivity of symbolic rules. Finally, temporal patterns of model reliance on skills were visualized using skill–timestep importance heatmaps, providing a global view of how the model's reasoning evolves.

\section{Results and Analysis}
\label{sec:results}

\subsection{Experimental setting}
\label{subsec:experimental_setting}

All neural-symbolic experiments were implemented using the PyNeuraLogic framework (Sec.~\ref{subsec:relational_nsai}). Experiments were run on a local workstation equipped with a 13th Gen Intel Core i7-1370P CPU and 32 GB RAM, with supplementary runs on Google Colab Pro+ for extended training. The neural backbone consists of a recurrent architecture (RNN) operating over symbolic embeddings (Sec.~\ref{subsec:architecture}). Skill, quiz, and correctness embeddings were each set to 16 dimensions, balancing the ability to capture meaningful interaction patterns with computational efficiency, and combined into a unified timestep representation. The recurrent core comprised two recurrent layers, followed by an explicit temporal shift grounded via symbolic $\text{next}(t-1, t)$ relations. Model outputs were produced using a sigmoid activation to estimate next-step correctness probabilities for the target quiz item at each timestep. Model training employed the Adam optimization algorithm with a learning rate of $1 \times 10^{-4}$, using binary cross-entropy as the objective function. Training proceeded for a maximum of 300 epochs, with early stopping triggered when the validation loss failed to improve for seven consecutive epochs. For the knowledge-augmented model, symbolic rules were assigned learnable weights, allowing the influence of each rule to be adjusted during training. The BaseNS-DKT retained the same architecture and training configuration but excluded the symbolic rule module, ensuring that any performance differences arise solely from the presence or absence of rule-based guidance. The Classic-DKT baseline was implemented in PyTorch using a purely neural recurrent architecture. To ensure a controlled comparison, the model used the same embedding dimensionality and recurrent structure as the other models. Each interaction was encoded using embeddings for skill, quiz, and correctness, combined through linear projections and processed by a two-layer RNN, with a sigmoid output layer predicting next-step correctness. The training configuration, including learning rate, binary cross-entropy loss, and early stopping strategy, was kept identical to the neural-symbolic models.

\subsection{Quantitative analysis of model performances}
\label{subsec:quantitative_results}

Table~\ref{tab:performance} presents the predictive performance of Responsible-DKT, BaseNS-DKT, and Classic-DKT across different sequence lengths and training ratios. Overall, the results consistently indicate that knowledge injection through interpretable rule-based guidance enhances both predictive quality and class balance under all experimental conditions. Regardless of sequence length or training ratio, Responsible-DKT outperforms the other models in terms of AUC, accuracy, and—most importantly—minority-class recall and F1-score. Across nearly all configurations, it achieves the highest AUC (up to 0.90) and accuracy (up to 0.86), with substantially improved recall and F1-scores for the low-performance class, which is typically more difficult to model due to class imbalance. Importantly, even in low-data settings the benefit of rule-based guidance is evident. With only 10\% of the training data and short sequences (length 10), NSAI with rules already achieves an AUC of 0.80, increasing to 0.88 for full sequence length. This corresponds to an improvement of up to 8 percentage points compared to the fully data-driven Classic-DKT, whose maximum AUC remains around 0.80 even with larger training ratios and longer sequences. Furthermore, performance in the rule-augmented model improves systematically as sequence length increases and as more training data become available, approaching 0.90 AUC. The BaseNS-DKT performs comparably to the Classic-DKT, with minor variations across configurations likely due to differences in training dynamics, possibly reflecting optimization behaviour (e.g., full-sequence vs. mini-batch training). However, in other configurations—such as 100\% training ratio with full sequences—it performs slightly below the data-driven baseline. This pattern suggests that while neural-symbolic structuring provides stability and competitive accuracy, the explicit injection of pedagogical rules is the key factor driving consistent and robust improvements across conditions. Collectively, these results demonstrate that rule-based knowledge injection does not merely refine performance marginally; rather, it systematically strengthens discrimination, improves minority-class behaviour, and enables more stable gains across both low-performing and high-performing settings.

\begin{table*}[ht]
\centering
\caption{Performance comparison of the models across varying sequence lengths and training ratios.}
\label{tab:performance}

\small
\begin{tabular}{llccccccccc}
\toprule
\multirow{3}{*}{\textbf{Models}} & 
\multirow{3}{*}{\textbf{\makecell{Training\\ratio (\%)}}} & 
\multirow{3}{*}{\textbf{N}} & 
\multicolumn{8}{c}{\textbf{Metrics}} \\
\cmidrule(lr){4-11}

& & & \multirow{2}{*}{\textbf{AUC}} 
& \multirow{2}{*}{\textbf{Accuracy}} 
& \multicolumn{2}{c}{\textbf{Precision}} 
& \multicolumn{2}{c}{\textbf{Recall}} 
& \multicolumn{2}{c}{\textbf{F1-score}} \\

\cmidrule(lr){6-7} \cmidrule(lr){8-9} \cmidrule(lr){10-11}
& & & & & \textbf{Low} & \textbf{High} & \textbf{Low} & \textbf{High} & \textbf{Low} & \textbf{High} \\
\midrule

\multirow{12}{*}{BaseNS-DKT}
& \multirow{4}{*}{10}
& 10   & 0.56 & 0.76 & 0.10 & 0.79 & 0.02 & 0.95 & 0.03 & 0.86 \\
& & 50   & 0.68 & 0.79 & 0.09 & 0.80 & 0.01 & 0.99 & 0.01 & 0.88 \\
& & 100  & 0.71 & 0.76 & 0.55 & 0.77 & 0.08 & 0.98 & 0.14 & 0.86 \\
& & Full & 0.77 & 0.82 & 0.62 & 0.84 & 0.27 & 0.96 & 0.38 & 0.89 \\

\cmidrule(lr){2-11}
& \multirow{4}{*}{50}
& 10   & 0.75 & 0.75 & 0.35 & 0.82 & 0.27 & 0.87 & 0.30 & 0.84 \\
& & 50   & 0.77 & 0.82 & 0.57 & 0.86 & 0.36 & 0.93 & 0.44 & 0.89 \\
& & 100  & 0.74 & 0.77 & 0.54 & 0.81 & 0.34 & 0.91 & 0.42 & 0.86 \\
& & Full & 0.78 & 0.82 & 0.56 & 0.87 & 0.45 & 0.91 & 0.50 & 0.89 \\

\cmidrule(lr){2-11}
& \multirow{4}{*}{100}
& 10   & 0.78 & 0.78 & 0.45 & 0.85 & 0.39 & 0.88 & 0.42 & 0.86 \\
& & 50   & 0.73 & 0.80 & 0.48 & 0.84 & 0.27 & 0.93 & 0.35 & 0.88 \\
& & 100  & 0.76 & 0.76 & 0.52 & 0.81 & 0.34 & 0.90 & 0.41 & 0.85 \\
& & Full & 0.78 & 0.82 & 0.56 & 0.88 & 0.53 & 0.89 & 0.55 & 0.89 \\

\midrule

\multirow{12}{*}{Responsible-DKT}
& \multirow{4}{*}{10}
& 10   & \textbf{0.80} & 0.78 & 0.43 & 0.81 & 0.18 & 0.94 & 0.26 & 0.87 \\
& & 50   & \textbf{0.85} & 0.84 & 0.66 & 0.86 & 0.36 & 0.95 & 0.47 & 0.90 \\
& & 100  & \textbf{0.87} & 0.82 & 0.71 & 0.84 & 0.46 & 0.94 & 0.56 & 0.89 \\
& & Full & \textbf{0.88} & 0.85 & 0.67 & 0.89 & 0.53 & 0.93 & 0.59 & 0.91 \\

\cmidrule(lr){2-11}
& \multirow{4}{*}{50}
& 10   & \textbf{0.86} & 0.83 & 0.60 & 0.87 & 0.49 & 0.91 & 0.54 & 0.89 \\
& & 50   & \textbf{0.87} & 0.86 & 0.68 & 0.89 & 0.55 & 0.94 & 0.61 & 0.91 \\
& & 100  & \textbf{0.88} & 0.84 & 0.72 & 0.87 & 0.58 & 0.93 & 0.65 & 0.90 \\
& & Full & \textbf{0.90} & 0.86 & 0.68 & 0.89 & 0.54 & 0.94 & 0.60 & 0.91 \\

\cmidrule(lr){2-11}
& \multirow{4}{*}{100}
& 10   & \textbf{0.86} & 0.85 & 0.63 & 0.90 & 0.63 & 0.90 & 0.63 & 0.90 \\
& & 50   & \textbf{0.88} & 0.86 & 0.68 & 0.90 & 0.59 & 0.93 & 0.63 & 0.92 \\
& & 100  & \textbf{0.88} & 0.84 & 0.70 & 0.87 & 0.58 & 0.92 & 0.63 & 0.90 \\
& & Full & \textbf{0.89} & 0.86 & 0.67 & 0.90 & 0.58 & 0.93 & 0.62 & 0.91 \\

\midrule

\multirow{12}{*}{Classic-DKT}
& \multirow{4}{*}{10}
& 10   & 0.73 & 0.84 & 0.66 & 0.87 & 0.47 & 0.94 & 0.55 & 0.90 \\
& & 50   & 0.73 & 0.83 & 0.69 & 0.84 & 0.27 & 0.97 & 0.39 & 0.90 \\
& & 100  & 0.74 & 0.78 & 0.61 & 0.80 & 0.27 & 0.94 & 0.37 & 0.87 \\
& & Full & 0.76 & 0.81 & 0.56 & 0.85 & 0.34 & 0.93 & 0.43 & 0.89 \\

\cmidrule(lr){2-11}
& \multirow{4}{*}{50}
& 10   & 0.75 & 0.83 & 0.63 & 0.88 & 0.51 & 0.92 & 0.56 & 0.90 \\
& & 50   & 0.75 & 0.83 & 0.61 & 0.87 & 0.45 & 0.93 & 0.52 & 0.90 \\
& & 100  & 0.75 & 0.80 & 0.62 & 0.84 & 0.47 & 0.91 & 0.54 & 0.87 \\
& & Full & 0.78 & 0.82 & 0.60 & 0.86 & 0.39 & 0.93 & 0.47 & 0.89 \\

\cmidrule(lr){2-11}
& \multirow{4}{*}{100}
& 10   & 0.75 & 0.82 & 0.58 & 0.86 & 0.45 & 0.91 & 0.51 & 0.89 \\
& & 50   & 0.76 & 0.84 & 0.66 & 0.87 & 0.45 & 0.94 & 0.54 & 0.91 \\
& & 100  & 0.76 & 0.81 & 0.64 & 0.85 & 0.49 & 0.91 & 0.55 & 0.88 \\
& & Full & 0.80 & 0.84 & 0.62 & 0.88 & 0.49 & 0.92 & 0.55 & 0.90 \\

\bottomrule
\end{tabular}
\end{table*}

Table~\ref{tab:stage_error} reports stage-wise error rates by dividing each student sequence into early, middle, and late segments, with values representing the percentage of misclassified instances in each stage. Responsible-DKT consistently achieves the lowest error rates across most sequence lengths, particularly in the early and middle phases, indicating more stable and temporally coherent predictions. This is especially important for adaptive learning systems, as early prediction errors can misguide personalization, leading to inappropriate difficulty adjustments, premature mastery assumptions, or unnecessary interventions. In contrast, the BaseNS-DKT shows consistently higher error rates, often performing worse than other models. Only in two late-stage cases (sequence lengths 10 and 50) does Classic-DKT achieve slightly lower late-stage error than the rule-based model. Given the pedagogical importance of accurate early and middle predictions, the overall pattern strongly supports the value of rule-based knowledge injection for improving temporal reliability and responsible adaptation.

\begin{table}[ht]
\centering
\caption{Early-, middle-, and late-stage prediction error rates (\%) across different sequence lengths.}
\label{tab:stage_error}
\begin{tabular}{llccc}
\toprule
Models & Sequence length & Early & Middle & Late \\
\midrule

\multirow{4}{*}{BaseNS-DKT}
 & 10   & 22.97 & 20.78 & 24.69 \\
 & 50   & 24.03 & 18.27 & 19.16 \\
 & 100  & 22.25 & 21.11 & 27.82 \\
 & Full & 21.15 & 16.12 & 16.56 \\

\multirow{4}{*}{Responsible-DKT}
 & 10   & \textbf{17.57} & \textbf{10.39} & 18.52 \\
 & 50   & \textbf{15.58} & \textbf{13.46} & 11.98 \\
 & 100  & \textbf{16.42} & \textbf{13.73} & \textbf{18.75} \\
 & Full & \textbf{14.35} & \textbf{15.28} & \textbf{13.54} \\

\multirow{4}{*}{Classic-DKT}
 & 10   & 22.97 & 16.88 & \textbf{16.05} \\
 & 50   & 21.10 & 15.71 & \textbf{10.48} \\
 & 100  & 19.33 & 18.65 & 19.96 \\
 & Full & 18.92 & 15.38 & 15.00 \\

\bottomrule
\end{tabular}
  \begin{minipage}{0.5\textwidth}
    \smallskip
    \footnotesize
    * All values are reported as percentages.
  \end{minipage}

\end{table}

Table~\ref{tab:volatility_inconsistency} reports volatility and inconsistency measures across models and sequence lengths. Overall, Responsible-DKT achieves the lowest inconsistency in all settings, indicating that its prediction updates are most consistently aligned with students' observed responses. Moreover, for the rule-based model, inconsistency decreases as sequence length increases, suggesting that longer learning histories further stabilize directional reliability. In contrast, both BaseNS-DKT and the Classic-DKT exhibit higher inconsistency values, reflecting less coherent alignment between prediction shifts and observed outcomes. At the same time, BaseNS-DKT shows the lowest volatility, followed closely by the Classic-DKT, whereas the rule-augmented model exhibits higher volatility. This pattern indicates that rule activation introduces more decisive and principled probability shifts when consistent evidence sequences occur (e.g., consecutive correct or incorrect responses), reflecting greater responsiveness to meaningful behavioural patterns rather than random fluctuation. By comparison, the lower volatility of the Classic-DKT and BaseNS-DKT models indicate smoother but possibly less responsive updates, which can contribute to their higher inconsistency rates.

\begin{table}[ht]
\centering
\caption{Volatility and inconsistency of predicted mastery trajectories across different sequence lengths.}
\label{tab:volatility_inconsistency}
\begin{tabular}{llcc}
\toprule
Models & Sequence length & Volatility & Inconsistency \\
\midrule

\multirow{4}{*}{BaseNS-DKT}
 & 10   & \textbf{0.12} & 0.48 \\
 & 50   & \textbf{0.09} & 0.46 \\
 & 100  & \textbf{0.11} & 0.43 \\
 & Full & \textbf{0.11} & 0.44 \\

\multirow{4}{*}{Responsible-DKT}
 & 10   & 0.18 & \textbf{0.41} \\
 & 50   & 0.17 & \textbf{0.40} \\
 & 100  & 0.17 & \textbf{0.36} \\
 & Full & 0.17 & \textbf{0.36} \\

\multirow{4}{*}{Classic-DKT}
 & 10   & 0.12 & 0.45 \\
 & 50   & 0.13 & 0.46 \\
 & 100  & 0.14 & 0.44 \\
 & Full & 0.13 & 0.44 \\

\bottomrule
\end{tabular}

\par\smallskip
\footnotesize
\centering
* Volatility and Inconsistency are reported as proportions (0--1 scale).

\end{table}

\subsection{Qualitative analysis of model performances: Sequential stability}
\label{subsec:qualitative_results}

\begin{figure}[t]
    \centering
    
    \begin{subfigure}{\linewidth}
        \includegraphics[width=\linewidth]{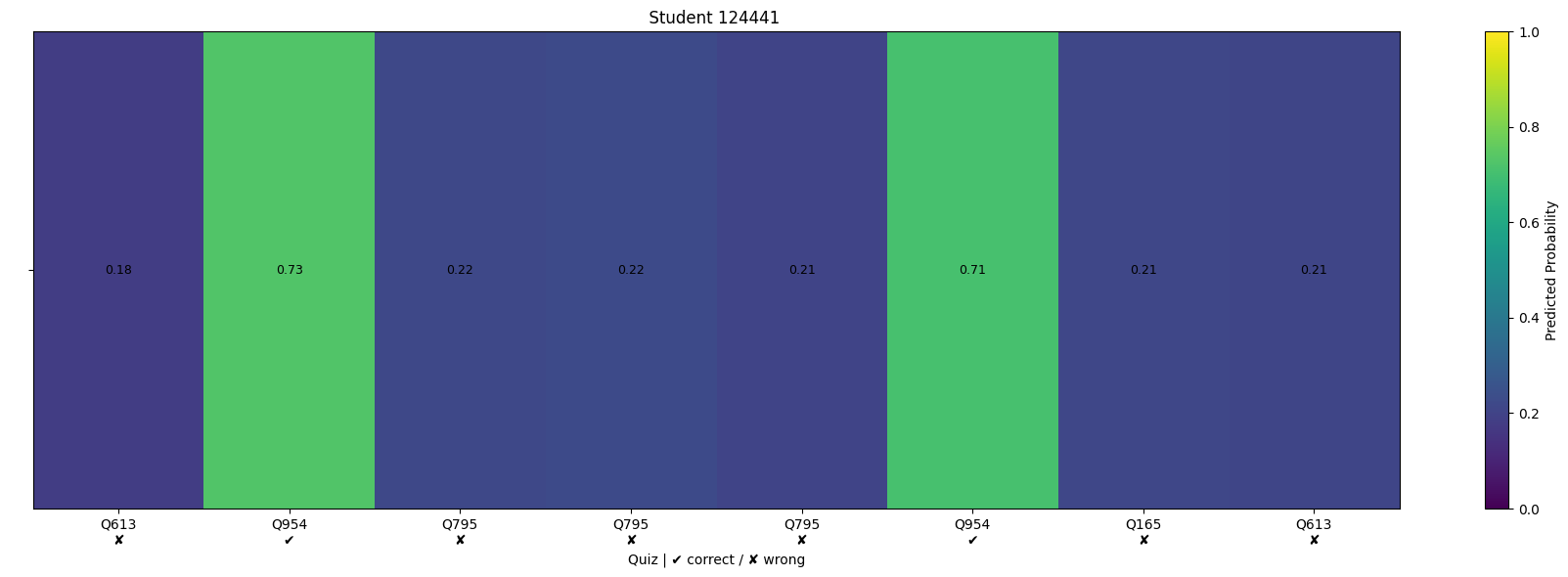}
        \caption{Responsible-DKT model}
        \label{fig:fig3a}
    \end{subfigure}
    
    \vspace{0.5cm} 
    
    \begin{subfigure}{\linewidth}
        \includegraphics[width=\linewidth]{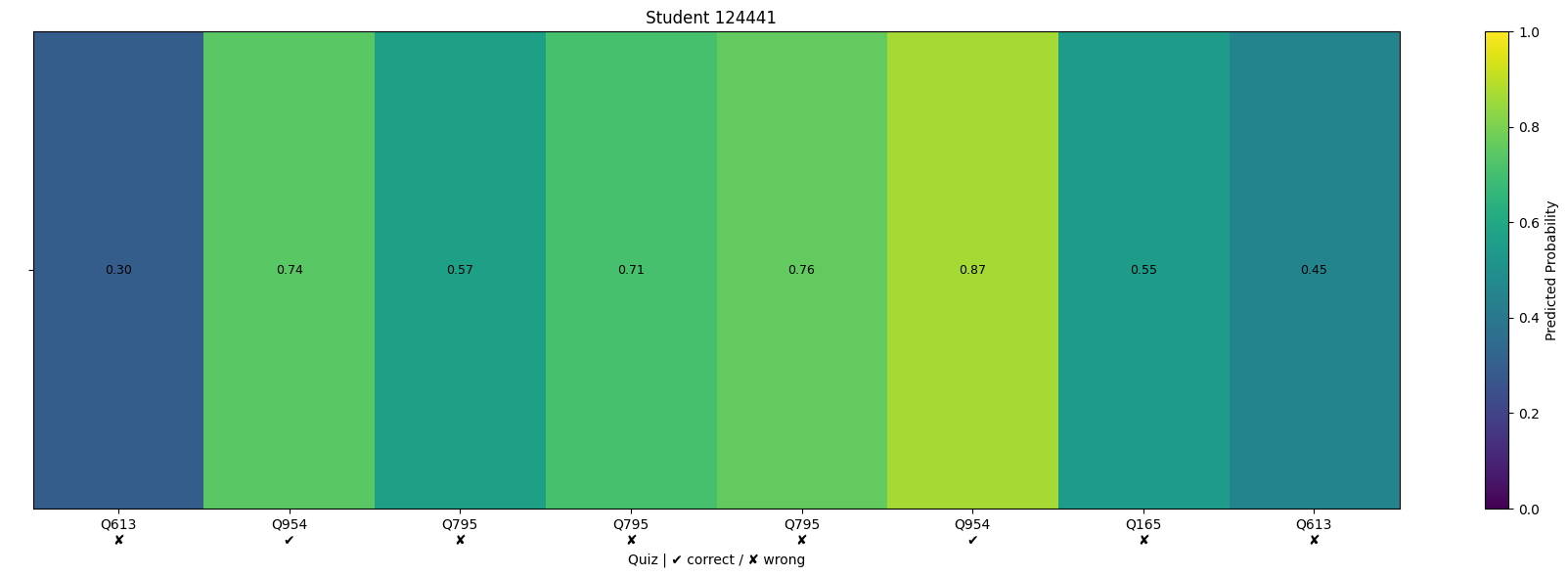}
        \caption{Classic-DKT model}
        \label{fig:fig3b}
    \end{subfigure}
    
    \caption{Step-wise predicted probabilities for repeated attempts of a single skill (\textit{Addition and subtraction of fractions}) by one student. Each column represents a quiz interaction involving the same skill, with $\checkmark$ indicating a correct response and $\times$ an incorrect response.}
    \label{fig:fig3}
\end{figure}

To illustrate rule-based behaviour within short sequences (10 interactions), the mastery and non-mastery rules were modified to activate after a single timestep rather than requiring multiple observations. Their weights were fixed rather than learned from data to clearly show their direct contribution to the predictions. Fig.~\ref{fig:fig3} shows that, for the skill \textit{Addition and subtraction of fractions}, the Responsible-DKT model updates predicted mastery more directly in response to observed student answers, whereas the Classic-DKT model exhibits smoother probability trajectories driven by hidden-state accumulation. Such smoothness can be desirable in knowledge tracing, as it reflects gradual updates of the learner's knowledge state. However, excessive smoothing may also lead to well-known issues in DKT models, particularly the reconstruction problem, where predicted mastery does not align with the observed student response \citep{krivich2025systematic, yeung2018addressing}. This can further result in temporal inconsistency, where mastery estimates change in a direction that contradicts the student's answer \citep{hooshyar2025problems}. At the first interaction, both models produce similar predictions. After the incorrect response at the third interaction, the Responsible-DKT model sharply decreases the predicted probability, reflecting the activation of the non-mastery rule. In contrast, the Classic-DKT model reduces the probability more moderately and subsequently increases it despite additional incorrect responses. When the student answers correctly at interaction six, the mastery rule activates and the prediction increases to 0.71, while the Classic-DKT model continues to rise to 0.87. After subsequent incorrect responses, the Responsible-DKT prediction drops sharply, whereas the Classic-DKT model decreases more gradually. Overall, these results suggest that incorporating symbolic mastery rules allows the Responsible-DKT model to regulate prediction updates more directly based on observed responses. While the neural component captures the gradual evolution of the learner's knowledge state, the injected rules help correct prediction updates when they become inconsistent with observed performance. The sharp probability changes in Responsible-DKT are largely due to the experimental setup, where rules were configured to fire after a single correct or incorrect response for illustration. In practice, rules can be defined over longer observation patterns, allowing the model to maintain gradual knowledge updates.

\subsection{Interpretability of the Responsible-DKT approach}
\label{subsec:interpretability_results}

\begin{figure}[t]
\centering
\input{figure/nesy_dkt_graph}
\caption{Grounded neural-symbolic computation graph for predicting $\text{correct}(t_1, q_{954})$. House nodes denote grounded input facts (quiz, skill, and prior correctness), which are first integrated into a unified representation via $\text{combined\_embed}(T)$, where quiz, skill, and outcome embeddings are merged. The yellow-highlighted nodes indicate the three main predictive pathways contributing to the target: the recurrent neural pathway (left), the symbolic mastery pathway (right), and the historical aggregation pathway (middle). These components are aggregated at the final output node to produce the predicted probability of correctness.}
\label{fig:fig4}
\end{figure}
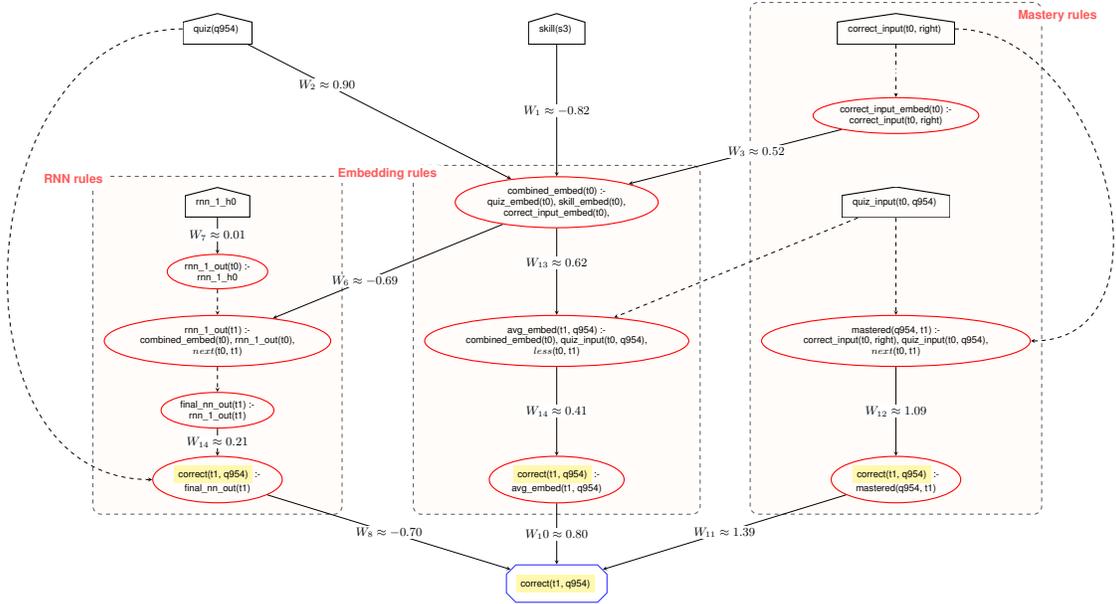

Because the Responsible-DKT model is formulated natively as a logic programming template, it bypasses the need for post-hoc, perturbation-based explainers (such as SHAP or LIME). Unlike conventional neural-based DKT models (e.g., Classic-DKT)—where the reasoning process remains hidden inside opaque matrix multiplications—Responsible-DKT's \textit{grounded computation graph} explicitly exposes the inferential structure behind each prediction. By tracking the exact gradients flowing backward from the target \textit{query} to the specific input \textit{facts} and pedagogical \textit{rules}, we can faithfully quantify the explicit relational reasoning path the model took. Each contribution—recurrent dynamics, symbolic mastery rules, and historical aggregation—can be traced as a distinct computational pathway with identifiable weights and transformations. This transparency allows researchers to inspect, validate, and even constrain the reasoning process, thereby making the Responsible-DKT model substantially more interpretable and aligned with pedagogical theory. Fig.~\ref{fig:fig4} illustrates the grounded computation graph for predicting $\text{correct}(t_1, q_{954})$, i.e., quiz 954 at a second timestep, in our Responsible-DKT model (see Sec.~\ref{subsubsec:responsible_dkt} for the respective model architecture and naming). For clarity and visual readability, the figure depicts a simplified configuration of the model, where the embedding dimensionality is reduced to 1, a single recurrent layer is used, and the symbolic rules are simplified to trigger mastery or non-mastery after a single correct or incorrect response. These adjustments were made solely to produce a more compact and interpretable visualization.

In Fig.~\ref{fig:fig4}, house-shaped nodes denote ``factual'' computation inputs, which represent grounded input facts to be embedded, including the quiz($q_{954}$), the $\text{skill}(s_3)$, and the binary correctness interaction fact $\text{correct\_input}(t_0, \text{right})$. Red elliptical nodes then correspond to grounded rules (Sec.~\ref{subsec:relational_nsai}), which implement differentiable conjunctive logic via learnable weights and activation functions. Finally, the blue node then corresponds to the specific target learning query, i.e., predicting correctness/mastery at the next (second) timestep $t_1$. The computational flow is then as follows. First, the symbolic inputs (quiz, skill, and correctness) are mapped into embedding representations and merged in a $\text{combined\_embed}(t_0)$ predicate via a weighted sum followed by a sigmoid transformation. The prediction is then composed from three interpretable pathways: (i) the $\text{combined\_embed}(t_0)$ representation is passed through a recurrent NN rule $\text{rnn\_1\_out}(t_1)$ (with tanh activation) together with the initial ($h_0$) hidden state, modelling the classic DKT temporal state evolution, (ii) a historical embedding aggregation pathway computing $\text{avg\_embed}(t_1, q_{954})$ over past interactions with the same quiz, and (iii) a symbolic mastery pathway where repeated correct responses instantiate $\text{mastered}(q_{954}, t_1)$ and influence the prediction through modulating the $\text{correct}(t_1, q_{954})$ predicate. The final blue query node then aggregates these components and applies a sigmoid to produce the probability of correctness. Solid edges denote weighted transformations, while dashed edges indicate purely symbolic information links. Together, the graph makes explicit how symbolic rules, embeddings, and temporal recurrence jointly contribute to the final prediction. Although the grounding process instantiates a distinct computation graph for each learner's unique interaction sequence at test time, a single set of embeddings and rule weights is learned globally during training, with quiz and skill embeddings, recurrent parameters, and rule weights being shared across all the learners, ensuring a principled generalization. Nevertheless, the activated inference structure, the particular hidden state evolution, and the contribution strengths of individual symbolic rules are uniquely determined by each learner's varying history. In this way, \textbf{Responsible-DKT framework provides highly individualized, explicitly traceable reasoning paths for every local prediction}.

\subsubsection{Local explanation}
\label{subsubsec:local_explanation}

\begin{figure}[t]
\centering
\includegraphics[width=\linewidth]{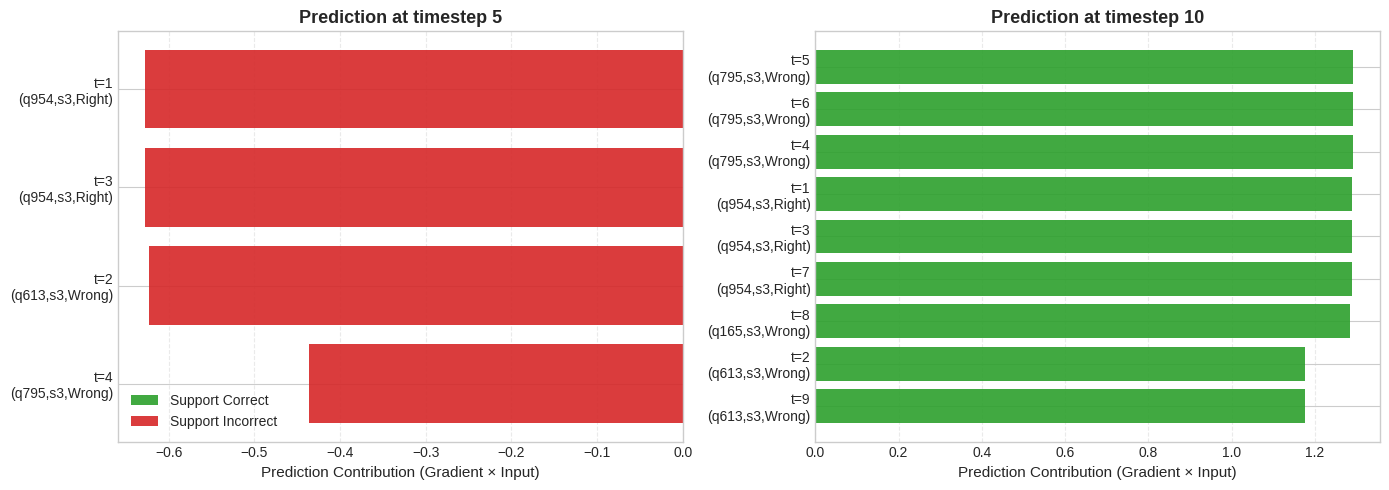}
\caption{Local explanation of Responsible-DKT predictions for a student sequence on skill $s_3$ (\textit{addition and subtraction of fractions}), showing contributions of previous interactions to predictions at timesteps 5 and 10 (left and right, respectively).}
\label{fig:fig5}
\end{figure}

To illustrate how the Responsible-DKT model makes predictions, we present a local explanation for a single test student with 10 interactions. The model predicts the correctness of the next response at each timestep using the previous interactions as context. Fig.~\ref{fig:fig5} shows explanations for predictions at timestep 5 and timestep 10. All interactions in this example correspond to skill $s_3$, which represents \textit{addition and subtraction of fractions}. The quizzes (e.g., $q_{954}$, $q_{613}$, $q_{795}$) are different exercises assessing this same skill. At timestep 5, previous interactions contribute negatively, pushing the model toward predicting an incorrect response, which matches the ground truth. The strongest negative contributions come from $t_1$ and $t_3$, followed by $t_2$, while $t_4$ contributes the least. These contributions are primarily shaped by the embedding-based aggregation ($\text{avg\_embed}$) of past interactions, while patterns of incorrect responses activate the $\text{not\_mastered}$ rule, reinforcing the negative prediction. At timestep 10, most prior interactions contribute positively and with similar strength, increasing the likelihood of a correct response, but the combined influence is insufficient to change the final prediction, resulting in a misclassification. Overall, $t_2$ consistently shows weaker influence across both predictions, which may indicate that this interaction provides less informative evidence about the student's mastery, while $t_4$ (for timestep 5) and $t_9$ (for timestep 10) have the smallest contributions, suggesting that the most recent interaction has the least influence on the prediction in this example.

\subsubsection{Global explanation}
\label{subsubsec:global_explanation}

\begin{figure}[t]
\centering
\includegraphics[width=\linewidth]{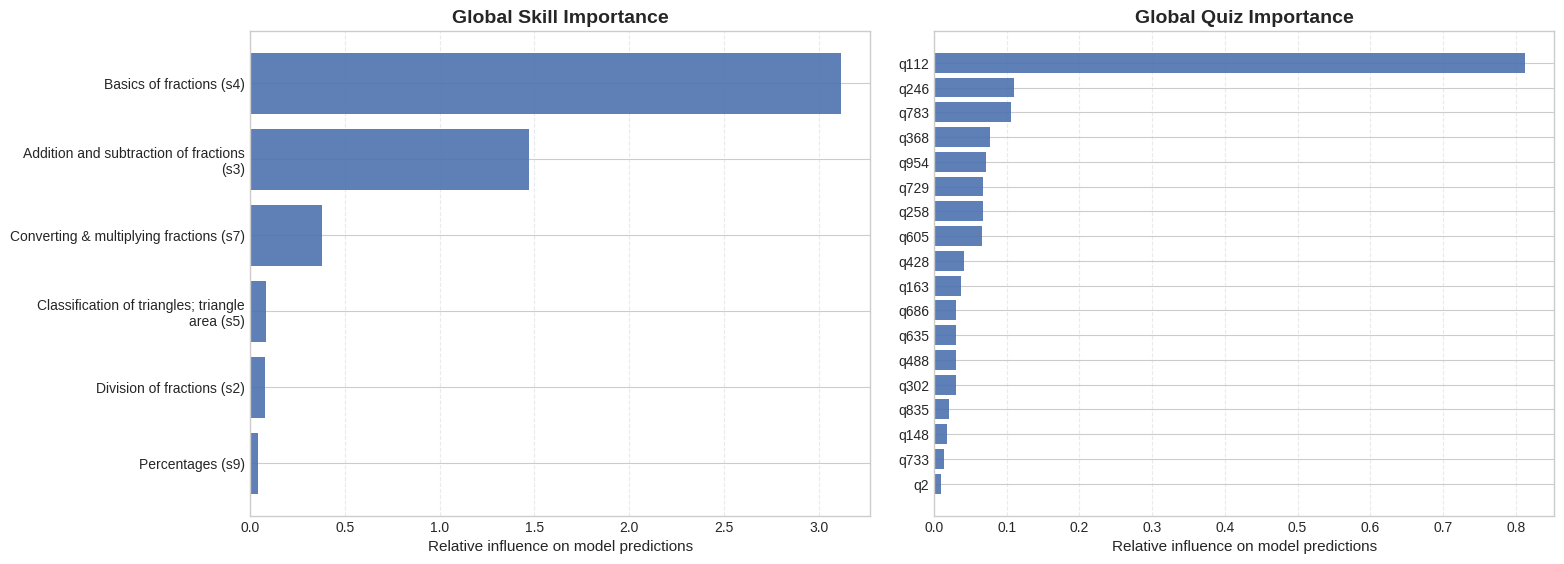}
\caption{Global explanation of the Responsible-DKT model predictions.}
\label{fig:fig6}
\end{figure}

Fig.~\ref{fig:fig6} presents the global explanation of the Responsible-DKT model for sequences of length 10, showing which skills and quizzes have the greatest influence on the model's predictions across the dataset. The left panel shows global skill importance, while the right panel shows global quiz importance. The y-axis lists the skills or quizzes, and the x-axis shows their relative influence on the model's predictions, computed by aggregating the gradient norms of input facts across all samples. Larger values indicate that changes in that input feature would have a stronger effect on the model's predictions. The results show that \textit{Basics of fractions} ($s_4$) has the strongest overall influence on the model, followed by \textit{Addition and subtraction of fractions} ($s_3$). \textit{Converting and multiplying fractions} ($s_7$) also contributes to the predictions but with substantially smaller influence, while the remaining skills have minimal impact. A similar pattern is observed for quizzes: a small number of exercises, particularly $q_{112}$, followed by $q_{246}$ and $q_{783}$, contribute more strongly to the model's predictions, whereas most quizzes have very low influence. Overall, the global explanation indicates that the model relies primarily on a small subset of fraction-related skills and a few specific exercises when forming predictions. This suggests that student performance on these concepts plays a central role in the model's assessment of knowledge, while other skills contribute comparatively less to the prediction process. These patterns are consistent with the rule-based analysis (see Table~\ref{tab:rule_importance}), where embedding-based aggregation dominates overall influence, while symbolic rules provide targeted adjustments when specific learning patterns, such as repeated errors, are detected.

To further examine how the injected symbolic knowledge influences the model's predictions, we analyse the global importance of the symbolic reasoning rules used within the Responsible-DKT model. Table~\ref{tab:rule_importance} presents the global importance of these rules across the dataset. The table reports three metrics for each rule aggregated over all samples: the average activation magnitude ($|\text{val}|$), the average gradient magnitude ($|\text{grad}|$), and the frequency of occurrence (count) in the computation graph. The activation magnitude is computed as the average absolute value of the rule neuron's output when the rule is instantiated, indicating \textit{how strongly the rule contributes to the forward computation}. The gradient magnitude is computed as the average absolute gradient of that neuron with respect to the loss, reflecting how sensitive the model's predictions are to changes in that rule. The count indicates \textit{how many times the rule appears across all samples in the computation graph}, showing how frequently the rule participates in the model's reasoning process. The results indicate that the $\text{avg\_embed}$ shows the highest activation and appears most frequently, indicating that the model primarily relies on the embedding-based aggregation of past interactions when forming predictions. In contrast, $\text{not\_mastered}$ appears much less often but has the largest gradient. Although its activation is slightly lower than $\text{avg\_embed}$, the larger gradient indicates that the model is more sensitive to this rule during learning, meaning that patterns of repeated errors contribute more strongly to parameter updates when they occur. Overall, this suggests that while the model mainly relies on embedding-based summaries of past interactions, it is particularly sensitive to patterns of repeated errors, indicating that non-mastery evidence plays a stronger role in shaping prediction updates. This also illustrates \textit{how neural-symbolic models can empirically examine pedagogical assumptions embedded in learner modelling rules}.

\begin{table}[t]
\centering
\caption{Global importance of symbolic rules in the Responsible-DKT model.}
\label{tab:rule_importance}
\begin{tabular}{lccc}
\toprule
\textbf{Rule} & \textbf{Avg $|\text{val}|^*$} & \textbf{Avg $|\text{grad}|^{**}$} & \textbf{Count} \\
\midrule
$\text{avg\_embed}$ & 1.9947 & 0.0008 & 284 \\
$\text{mastered}$ & 1.6692 & 0.0004 & 219 \\
$\text{not\_mastered}$ & 1.8557 & 0.0016 & 49 \\
\bottomrule
\multicolumn{4}{l}{\small $^*|\text{val}|$: average absolute rule activation (strength of contribution during forward computation).} \\
\multicolumn{4}{l}{\small $^{**}|\text{grad}|$: average absolute gradient (sensitivity of model predictions to that rule).}
\end{tabular}
\end{table}

\section{Discussion and Conclusions}
\label{sec:discussion}

The objective of this study was to investigate whether neural-symbolic AI can support more responsible learner modelling by integrating structured pedagogical knowledge into deep knowledge tracing. To this end, we proposed Responsible-DKT, a neural-symbolic approach that embeds symbolic educational knowledge in the form of simple rules inspired by empirically derived pedagogical patterns \citep[e.g.,][]{koedinger2023astonishing} into sequential neural architectures using the PyNeuraLogic framework. The study addressed three research questions concerning predictive performance and generalizability (RQ1), sequential stability and learning dynamics (RQ2), and interpretability of model predictions (RQ3).

\subsection{Interpretation and comparison of the results}

Regarding \textit{RQ1}, the results (Sec \ref{subsec:quantitative_results}) show that integrating symbolic knowledge can improve predictive performance and robustness compared with purely data-driven approaches. Across different sequence lengths and training ratios, Responsible-DKT consistently achieved higher AUC and accuracy than the Classic-DKT baseline, while also improving recall and F1-score for the minority class. Importantly, the model maintained strong performance even with limited training data, achieving more than 0.80 AUC with only a small fraction of the training set. This suggests that \textbf{incorporating domain knowledge can compensate for data limitations and reduce the dependence on large-scale datasets, which are often difficult to obtain in educational contexts}. This finding aligns with those reported by \citet{hooshyar2024augmenting, hooshyar2025towardsb}, who showed that neural-symbolic approaches integrating educational knowledge can improve model generalizability and robustness compared with purely data-driven neural networks. Although their study focused on non-sequential data and used a different neural-symbolic framework based on latent variables embedded within the network structure, the results similarly suggest that knowledge injection—both in sequential and non-sequential settings—can enhance model performance. This also resonates with the work of \citet{shakya2021student}, who integrated symbolic educational relationships with LSTM models using Markov Logic and trained the network on representative samples rather than the entire dataset. Their approach improved training efficiency and reduced overfitting, further illustrating how combining symbolic knowledge with neural models can mitigate data-related limitations in educational AI. Similarly, \citet{tato2022infusing} proposed a hybrid architecture combining deep neural networks with expert knowledge through an attention mechanism to address challenges such as data imbalance and limited training data in learner modelling. Their results demonstrated improved prediction of students' knowledge states and learning outcomes, highlighting the value of integrating expert knowledge into neural models to support generalization when educational datasets are small or imbalanced. The present study extends this line of work by demonstrating that these benefits also hold in sequential knowledge tracing settings, where models must learn from temporally ordered interaction data. In particular, our results show that symbolic knowledge injection can improve predictive performance and robustness across varying sequence lengths and training ratios, while maintaining strong performance even under limited data conditions. Together, these studies reinforce the potential of knowledge-informed neural approaches for educational AI, where limited and uneven data availability remains a common challenge.

With respect to \textit{RQ2}, the results (Sec \ref{subsec:quantitative_results} and \ref{subsec:qualitative_results}) indicate that symbolic knowledge injection contributes to more stable and pedagogically coherent prediction dynamics. Responsible-DKT produced \textit{lower inconsistency rates and more reliable early-stage predictions compared with the baseline models}. This is particularly important for adaptive learning systems, where inaccurate early predictions can lead to inappropriate instructional decisions. The analysis also showed that rule activation enables the model to respond more directly to meaningful behavioural patterns, such as repeated correct or incorrect responses, improving alignment between prediction updates and observed student performance. Although this responsiveness can produce sharper probability shifts than those observed in purely neural models, these adjustments reflect principled reactions to evidence rather than arbitrary fluctuations. These findings also contribute to the broader discussion of temporal reliability in learner modelling, an aspect that remains largely overlooked in the literature. As highlighted in the systematic review by \citet{krivich2025systematic}, most studies evaluate DKT models using standard classification metrics (e.g., AUC or accuracy) while rarely assessing sequential stability or prediction consistency over time. By explicitly analysing early, middle, and late-stage errors, as well as inconsistency and volatility measures, our study provides additional insight into the temporal reliability of learner modelling systems. This issue has been previously noted by \citet{yeung2018addressing}, who showed that DKT models may produce inconsistent mastery trajectories and even decrease predicted mastery after correct responses. They addressed this by adding regularization terms to the loss function to smooth prediction transitions. However, such modifications remain purely statistical and do not allow for the incorporation of pedagogical knowledge, the inspection of the decision-making process, or safeguards against learning spurious correlations and biases. In contrast, our approach regulates prediction dynamics through symbolic knowledge injection, enabling the model to encode educational principles and provide interpretable reasoning for prediction updates. Recent work by \citet{hooshyar2025problems} further emphasizes the importance of such analyses. In their case study comparing data-driven DKT models with an LLM, DKT achieved higher predictive performance and demonstrated substantially stronger temporal coherence, while the LLM-based approach produced inconsistent mastery trajectories and incorrect directional updates over time—even after extensive fine-tuning. In the present study, we show that temporal reliability can be further improved by integrating symbolic knowledge within neural architectures. The results suggest that neural-symbolic learner models can strengthen both predictive accuracy and temporal consistency, which are critical properties for trustworthy adaptive educational systems.

Beyond performance improvements, the proposed approach also demonstrates an important methodological contribution: \textbf{it allows stakeholders to explicitly inject pedagogical knowledge into the learner modelling process}. In the current study, simple mastery and non-mastery rules were introduced to guide prediction updates. However, the neural-symbolic framework supports a much broader range of knowledge integration mechanisms. For example, further rules could model spacing effects based on time gaps between interactions, or recency/weighted averages of past successes on the same skill or quiz. Other rules could capture transitions in learner behaviour—such as shifts from correct to incorrect answers or vice versa—which may signal uncertainty, forgetting, or guessing \citep{abdelrahman2023knowledge}. More structurally, symbolic constraints may also be derived from domain knowledge graphs, including prerequisite relationships between skills or relative difficulty levels, enabling transparent and pedagogically meaningful mastery propagation grounded in expert knowledge \citep[see][]{koedinger2023astonishing}. Finally, \textbf{beyond explicit rule-based influences, the framework also enables the introduction of latent intermediate representations}, analogous in spirit to KBANN-style knowledge injection \citep{hooshyar2024augmenting, towell1994knowledge}. Such latent variables can be structurally positioned between observed interaction inputs and prediction targets, encouraging the model to reason through meaningful learner states rather than relying solely on surface-level correlations. Examples include latent constructs such as learning momentum, engagement state, or fatigue, which capture gradual changes in learner behaviour and modulate predictions in a transparent yet flexible manner. In this way, neural-symbolic modelling provides a flexible mechanism for embedding educational theory into data-driven learning systems.

Finally, addressing \textit{RQ3}, the Responsible-DKT model provides interpretable explanations of predictions through its explicit neural-symbolic computation structure (Sec \ref{subsec:interpretability_results}). Unlike conventional DKT models, where reasoning remains hidden within opaque neural states, the grounded computation graph exposes how embeddings, symbolic rules, and recurrent dynamics jointly contribute to each prediction. This structure enables \textbf{explanation at multiple levels}. Local explanations reveal how previous interactions influence individual predictions over time, while global analyses identify which skills, quizzes, and symbolic rules most strongly affect model behaviour across the dataset. The results show that the model primarily relies on a small subset of key skills and exercises when forming predictions, while symbolic rules—particularly those capturing repeated incorrect responses—play an important role in regulating prediction updates. These analyses demonstrate that the model's reasoning process can be inspected and traced through explicit computational pathways, allowing researchers and educators to better understand how predictions are produced. This extends prior work by providing intrinsic, mechanism-level interpretability within a sequential knowledge tracing setting, where explanations are directly grounded in the model’s computation rather than inferred through post-hoc techniques \citep{bai2024survey}. Such interpretability is particularly important given that it remains largely overlooked in the knowledge tracing literature. For example, the recent systematic review by \citet{krivich2025systematic} shows that most studies do not explicitly address interpretability, and the few that do typically rely on post-hoc explanation techniques such as SHAP, attention-weight analysis, or Grad-CAM. While these approaches can highlight influential inputs, they do not necessarily reflect the true reasoning process of the model and may lead to misleading interpretations \citep{hooshyar2024problems, slack2020fooling}. Moreover, many studies equate visualizations of predicted skill mastery probabilities (i.e., heatmaps) with interpretability, even though such summaries do not reveal the mechanisms driving predictions \citep{krivich2025systematic, li2023calibrated}. As argued by \citet{rudin2019stop}, relying on post-hoc explanations for opaque models in high-stakes domains can be problematic, and designing models that are inherently interpretable offers a more reliable alternative. In line with this view, the neural-symbolic design adopted in Responsible-DKT provides intrinsic interpretability, allowing the decision-making process to be examined directly through its computational structure and reducing the risk of relying on spurious correlations or hidden biases \citep{hooshyar2024augmenting, hooshyar2024problems, rudin2019stop, tato2022infusing}. Moreover, such transparency can support \textbf{hypothesis testing and theory refinement}. By analysing the behaviour and importance of injected rules, one can evaluate whether the encoded pedagogical assumptions hold in practice—such as revealing that non-mastery patterns play a stronger role in prediction updates than mastery patterns—creating a feedback loop in which empirical evidence informs the refinement of both the model and the underlying learning theory. This represents a step beyond explanation toward theory-informed modelling, where interpretability is used not only for inspection but also for validating and refining educational assumptions. This is particularly important given that there are currently very few examples in the AI in education literature of studies that systematically test and refine learning theories at scale through computational modelling approaches \citep{giannakos2023role}.

Overall, the findings of the current study suggest that neural-symbolic approaches offer a promising pathway toward more responsible educational AI. By combining the pattern-learning capacity of neural networks with the transparency and structured reasoning of symbolic knowledge, Responsible-DKT demonstrates how learner models can become more interpretable, stable, and aligned with pedagogical principles. Beyond improving predictive performance, such hybrid human–AI approaches enable educators and researchers to contribute domain knowledge directly to the modelling process, supporting human-centred and theory-informed AI design. These characteristics are essential for the reliable deployment of AI-driven tutoring systems in real educational settings, where transparency, trust, and pedagogically grounded decisions are as important as predictive accuracy.

\subsection{Limitations and future work}
\label{subsec:limitations}

While the findings provide useful insights, several limitations remain that should be addressed in future research. First, the symbolic rules used in this study are relatively simple and do not fully reflect the broader range of capabilities offered by neural-symbolic approaches. Future work should explore more complex rule structures and richer forms of domain knowledge (e.g., mental models of biological systems) to better capture the full potential of neural-symbolic integration in learner modelling. Second, the experiments relied on a single dataset and a specific knowledge tracing task formulation. Evaluating the proposed approach across diverse datasets, learner modelling tasks, and educational contexts would help assess the generalisability of the findings and support responsible deployment in real educational settings such as K–12 environments. Third, the experimental comparison focused mainly on the classic DKT baseline and a neural-symbolic variant without rule injection. Future research should therefore compare the proposed approach with a broader range of state-of-the-art knowledge tracing models, including attention-based and graph-based architectures. Finally, although this study is framed within the broader perspective of responsible AI, the current implementation should be understood as a partial and methodological contribution rather than a comprehensive realisation of all responsible AI dimensions. The injected rules are pedagogically motivated and do not directly operationalise ethical principles, fairness constraints, or privacy-preserving mechanisms. Instead, they demonstrate how symbolic knowledge can be embedded into neural architectures to support human-centred design, by enabling the explicit incorporation of expert pedagogical knowledge; transparent decision-making, through explicit reasoning structures, learned rule weights, and interpretable computation graphs; and more reliable and ethically grounded predictions, as rule injection acts as a structural constraint that reduces sequential instability and mitigates reliance on spurious correlations. Therefore, the notion of responsibility considered here is primarily grounded in pedagogical aspects of learner modelling, rather than the full, broader scope of responsible AI. Future work should extend this framework by incorporating richer forms of knowledge, including fairness-aware constraints, causal relationships, and privacy-preserving inductive biases, as well as by empirically evaluating their impact on user trust, ethical decision-making, and real-world educational deployment.
\section*{Acknowledgments}
This work was supported by the Estonian Research Council grant (PRG2215). G.\v{S}. acknowledges support from the Czech Science Foundation grant no.~26-22501S. The work of Dragan Ga\v{s}evi\'c was partially supported by the Australian Research Council (DP220101209) and the Jacobs Foundation (CELLA2CERES). We also thank the Opiq learning environment for their collaboration in this study.
\bibliographystyle{plainnat}
\bibliography{references}

\end{document}

\bibliographystyle{unsrtnat}
\bibliography{references}  






\end{document}

%% file: figure/nesy_dkt_workflow.tex
\resizebox{\textwidth}{!}{%
\begin{tikzpicture}[
    x=1cm, y=1cm,
    >=stealth,
    every node/.style={font=\sffamily\scriptsize, text=darktext},
    every path/.style={line cap=round, line join=round},
    phasebox/.style={draw=bordergray, thick, dashed, rounded corners=12pt, fill=lightbg},
    phasetitle/.style={font=\sffamily\bfseries\large, text=darktext},
    basebox/.style={draw=darktext, thick, rounded corners=4pt, fill=white, align=center, inner sep=8pt},
    flowarrow/.style={->, thick, darktext},
    neuralarrow/.style={->, thick, primary},
    symbolicarrow/.style={->, thick, accent},
    jointarrow/.style={->, line width=1.5pt, draw=secondary},
    gradarrow/.style={<-, thick, red}
]

\fill[phasebox] (-1, 2.5) rectangle (9.5, -4.8);
\node[phasetitle, anchor=north] at (4.25, 2.5) {1. Data Pre-processing};
\begin{scope}[yshift=-0.6cm, xshift=-0.3cm, xscale=0.92, transform shape]
\node[cylinder, draw=darktext, thick, fill=white, aspect=0.3, minimum height=1.5cm, minimum width=1.4cm, shape border rotate=90, align=center, font=\sffamily\small] (orig) at (0.5, 0.8) {Opiq\\Logs};

\node[font=\sffamily\bfseries, text=darktext] at (4.0, 1.75) {Filtering};
\begin{scope}[shift={(4.0, 0.75)}, rotate=90, scale=0.75]
    \draw[thick, darktext, fill=bordergray!30] (-0.6, 0.5) -- (0.6, 0.5) -- (0.15, -0.2) -- (0.15, -0.8) -- (-0.15, -0.8) -- (-0.15, -0.2) -- cycle;

    \fill[secondary] (-0.3, 0.8) circle (3pt);
    \fill[accent] (0.2, 0.9) circle (3pt);
    \draw[accent, thick] (0.05, 0.75) -- (0.35, 1.05);
    \draw[accent, thick] (0.05, 1.05) -- (0.35, 0.75); 
    \fill[secondary] (0.1, 1.2) circle (3pt);
    \fill[accent] (-0.2, 1.4) circle (3pt);
    \draw[accent, thick] (-0.35, 1.25) -- (-0.05, 1.55);
    \draw[accent, thick] (-0.35, 1.55) -- (-0.05, 1.25); 

    \fill[secondary] (0, -1.1) circle (3pt);
    \fill[secondary] (0, -1.5) circle (3pt);
\end{scope}

\begin{scope}[shift={(8.0, 0.8)}]
    \fill[white, draw=bordergray, dashed, thick, rounded corners=4pt] (-2.0, -1.0) rectangle (2.0, 1.0);
    
    \node[draw=darktext, fill=white, rounded corners=2pt, font=\ttfamily\scriptsize, inner sep=3pt] (cat1) at (-1.0, 0.4) {"Algebra"};
    \node[draw=darktext, fill=white, rounded corners=2pt, font=\ttfamily\scriptsize, inner sep=3pt] (cat2) at (-1.0, -0.4) {"Right"};
    
    \node[draw=primary, fill=primary!10, rounded corners=2pt, font=\ttfamily\scriptsize, inner sep=3pt] (id1) at (1.0, 0.4) {ID: 42};
    \node[draw=primary, fill=primary!10, rounded corners=2pt, font=\ttfamily\scriptsize, inner sep=3pt] (id2) at (1.0, -0.4) {y: 1};
    
    \draw[flowarrow] (cat1) -- (id1);
    \draw[flowarrow] (cat2) -- (id2);
\end{scope}

\draw[flowarrow] (1.3, 0.8) -- (2.6, 0.8);
\draw[flowarrow] (5.3, 0.8) -- (5.8, 0.8);


\begin{scope}[shift={(0.5, -3.5)}]
    \fill[darktext] (0, 0.15) circle (0.12);
    \draw[darktext, thick, rounded corners=1.5pt, fill=darktext!20] (-0.25, -0.15) -- (-0.25, -0.05) .. controls (-0.25, 0.05) and (0.25, 0.05) .. (0.25, -0.05) -- (0.25, -0.15) -- cycle;
\end{scope}

\node[font=\sffamily\bfseries, text=darktext] at (4.75, -.5) {Chronological Sequences};
\begin{scope}[shift={(2.0, -1.2)}]
    \node[draw=primary, thick, fill=primary!10, rounded corners=2pt, minimum height=0.6cm, minimum width=0.6cm] (x1) at (0,0) {$x_1$};
    \node[draw=primary, thick, fill=primary!10, rounded corners=2pt, minimum height=0.6cm, minimum width=0.6cm] (x2) at (1.5,0) {$x_2$};
    \node[draw=primary, thick, fill=primary!10, rounded corners=2pt, minimum height=0.6cm, minimum width=0.6cm] (x3) at (3,0) {$x_3$};
    \node (dots) at (4.2,0) {$\dots$};
    \node[draw=primary, thick, fill=primary!10, rounded corners=2pt, minimum height=0.6cm, minimum width=0.6cm] (xt) at (5.5,0) {$x_n$};
    
    \node[font=\sffamily\scriptsize, text=darktext, align=center] at (0,-0.8) {$(s_1, q_1, y_1)$};
    \node[font=\sffamily\scriptsize, text=darktext, align=center] at (1.5,-0.8) {$(s_2, q_2, y_2)$};
    \node[font=\sffamily\scriptsize, text=darktext, align=center] at (3,-0.8) {$(s_3, q_3, y_3)$};
    \node[font=\sffamily\scriptsize, text=darktext, align=center] at (5.5,-0.8) {$(s_n, q_n, y_n)$};

    \draw[flowarrow] (x1) -- (x2);
    \draw[flowarrow] (x2) -- (x3);
    \draw[flowarrow] (x3) -- (dots);
    \draw[flowarrow] (dots) -- (xt);
\end{scope}

\node[font=\sffamily\bfseries, text=darktext] at (4.75, -2.8) {Student-level Partitioning};
\begin{scope}[shift={(1.25, -3.8)}]
    \fill[primary!70] (0,0) rectangle (4.5, 0.6);
    \fill[secondary!70] (4.5,0) rectangle (5.5, 0.6);
    \fill[accent!70] (5.5,0) rectangle (7, 0.6);
    \draw[darktext, thick] (0,0) rectangle (7, 0.6);
    \draw[darktext, thick] (4.5,0) -- (4.5,0.6);
    \draw[darktext, thick] (5.5,0) -- (5.5,0.6);
    
    \node[font=\sffamily\small\bfseries, text=white] at (2.25, 0.3) {Training};
    \node[font=\sffamily\small\bfseries, text=white] at (5.0, 0.3) {Val};
    \node[font=\sffamily\small\bfseries, text=white] at (6.25, 0.3) {Test};
\end{scope}

\end{scope}

\fill[phasebox] (10.0, 2.5) rectangle (21, -4.8);
\node[phasetitle, anchor=north] at (15.5, 2.5) {2. Symbolic Sample Generation};


\begin{scope}[shift={(11.1, 0.7)}, yscale=1.3]
    \node[draw=darktext, thick, fill=white, rounded corners=2pt, minimum width=0.85cm, minimum height=0.6cm] (xt2) at (0, 0) {$x_{t-2}$};
    \node[draw=darktext, thick, fill=white, rounded corners=2pt, minimum width=0.85cm, minimum height=0.6cm] (xt1) at (2.6, 0) {$x_{t-1}$};
    \node[draw=darktext, thick, fill=white, rounded corners=2pt, minimum width=0.85cm, minimum height=0.6cm] (xt)  at (5.2, 0) {$x_{t}$};
    \node[draw=accent, thick, fill=accent!10, rounded corners=2pt, minimum width=1.45cm, minimum height=0.85cm, align=center, font=\sffamily\bfseries\scriptsize, text=accent] (yt1) at (8.4, 0) {Target\\$\hat{y}_{t+1}$};

    \draw[flowarrow, primary, dashed] (xt2) -- (xt1)
        node[midway, above=10pt, font=\ttfamily\scriptsize, text=primary] {next(t-2, t-1)};
    \draw[flowarrow, primary, dashed] (xt1) -- (xt)
        node[midway, above=10pt, font=\ttfamily\scriptsize, text=primary] {next(t-1, t)};
    \draw[flowarrow, primary, dashed] (xt) -- (yt1)
        node[midway, above=10pt, xshift=0pt, font=\ttfamily\scriptsize, text=primary] {next(t, t+1)};

    \draw[flowarrow] (xt) -- (5.2, -1.0) node[midway, right, font=\sffamily\scriptsize] {$(s_t, q_t, y_t)$};

    \node[draw=darktext, thick, fill=white, rounded corners=2pt, font=\ttfamily\scriptsize, inner sep=3pt] (f1) at (5.2, -1.4) {skill\_input(t, st)};
    \node[draw=darktext, thick, fill=white, rounded corners=2pt, font=\ttfamily\scriptsize, inner sep=3pt] (f2) at (5.2, -2.1) {quiz\_input(t, qt)};
    \node[draw=darktext, thick, fill=white, rounded corners=2pt, font=\ttfamily\scriptsize, inner sep=3pt] (f3) at (5.2, -2.8) {correct\_input(t, yt)};

    \draw[->, thick, dashed, bordergray] (xt1) -- (2.6, -1.0);
    \node[draw=bordergray, dashed, thick, fill=white, rounded corners=2pt, font=\ttfamily\scriptsize, inner sep=3pt, text=gray, align=center] (fx1) at (2.6, -2.1) {Ground Facts\\(t-1)};
    \draw[->, thick, dashed, bordergray] (xt2) -- (0, -1.0);
    \node[draw=bordergray, dashed, thick, fill=white, rounded corners=2pt, font=\ttfamily\scriptsize, inner sep=3pt, text=gray, align=center] (fx2) at (0, -2.1) {Ground Facts\\(t-2)};

    \draw[flowarrow, accent, line width=1pt] (yt1) -- (8.4, -1.0);
    \node[draw=accent, thick, fill=white, rounded corners=2pt, font=\ttfamily\scriptsize\bfseries, text=accent, inner sep=4pt] (query) at (8.4, -2.1) {? correct(t+1, q)};

    \draw[decorate,decoration={brace,amplitude=5pt,mirror},thick, darktext] (-.5,-3.4) -- (6.7,-3.4)
        node[midway,below=5pt, font=\sffamily\scriptsize\bfseries] {Context Base (Logical Ground Facts)};
    \draw[decorate,decoration={brace,amplitude=5pt,mirror},thick, accent] (6.8,-3.4) -- (9.8,-3.4)
        node[midway,below=5pt, font=\sffamily\scriptsize\bfseries] {Logical Query};

\end{scope}

\draw[flowarrow, line width=1.5pt] (9.5, -1.5) -- (10.0, -1.5); 

\fill[phasebox] (-1, -5.5) rectangle (21, -13);
\node[phasetitle, anchor=north] at (10, -5.5) {3. Neural-Symbolic Model Architecture};

\node[font=\sffamily\bfseries, text=darktext] at (3.5, -6.6) {Differentiable Logic Program};
\begin{scope}[shift={(-0.5, -12)}]
    \fill[white, draw=bordergray, thick, rounded corners=6pt] (0, -0.5) rectangle (7.5, 5.2);
    
    \node[anchor=west, font=\ttfamily\scriptsize, text=primary] at (0.2, 4.8) {\% Neural Model Rules};
    \node[anchor=west, font=\ttfamily\small] at (0.2, 4.4) {h(T) $\leftarrow$ rnn\_step( h(T-1), z(T) )};
    \node[anchor=west, font=\ttfamily\small] at (0.2, 4.0) {\dots};
    
    \node[anchor=west, font=\ttfamily\scriptsize, text=accent] at (0.2, 3.6) {\% Symbolic Rules};
    \node[anchor=west, font=\ttfamily\small] at (0.2, 3.2) {mastered(T) $\leftarrow$ correct(T-1) $\wedge$ correct(T-2)};
    \node[anchor=west, font=\ttfamily\small] at (0.2, 2.7) {avg\_emb(T) $\leftarrow$ mean\{ z(1..T) \}};
    \node[anchor=west, font=\ttfamily\small] at (0.2, 2.2) {\dots};
    
    \node[anchor=west, font=\ttfamily\scriptsize, text=secondary] at (0.2, 1.8) {\% Joint Interaction};
    \node[anchor=west, font=\ttfamily\small] at (0.2, .7) {$%
        \begin{aligned}
            \text{predict}(T{+}1) \leftarrow {} &\; h(T) ~\oplus\\
            &\; \text{mastered}(T) ~\oplus\\
            &\; \text{avg\_emb}(T) ~\oplus\\
            &\; \dots
        \end{aligned}
    $};
\end{scope}

\draw[flowarrow, line width=1.5pt] (7.2, -9.5) -- (8.8, -9.5) node[midway, above, font=\sffamily\small\bfseries] {Grounding};
\node[font=\sffamily\scriptsize, text=darktext, align=center] at (8.0, -10.2) {Unrolling via\\ground facts};

\node[font=\sffamily\bfseries, text=darktext] at (14.5, -6.6) {Neural Computation Graph};
\begin{scope}[shift={(9, -12)}]
    \fill[white, draw=bordergray, thick, rounded corners=6pt] (0, -0.5) rectangle (11.5, 5.2);
    
    \node[draw=darktext, fill=white, thick, rounded corners=2pt, minimum width=0.8cm, align=center] (x1) at (2, 0.1) {$x_{t-2}$};
    \node[draw=darktext, fill=white, thick, rounded corners=2pt, minimum width=0.8cm, align=center] (x2) at (5, 0.1) {$x_{t-1}$};
    \node[draw=darktext, fill=white, thick, rounded corners=2pt, minimum width=0.8cm, align=center] (x3) at (8, 0.1) {$x_{t}$};
    
    \foreach \xpos/\t in {2/t-2, 5/t-1, 8/t} {
        \draw[flowarrow] (\xpos, 0.5) -- (\xpos, 1.0);
        \fill[primary!40, draw=darktext, thin] (\xpos-0.4, 1.0) rectangle (\xpos+0.4, 1.2);
        \fill[secondary!40, draw=darktext, thin] (\xpos-0.4, 1.2) rectangle (\xpos+0.4, 1.4);
        \fill[accent!40, draw=darktext, thin] (\xpos-0.4, 1.4) rectangle (\xpos+0.4, 1.6);
        \node[font=\sffamily\scriptsize, anchor=west] at (\xpos+0.4, 1.3) {$z_{\t}$};
    }
    
    \node[font=\sffamily\tiny, align=left, anchor=west] at (9.5, 1.5) {\textbf{Embeddings}\\ \textcolor{primary}{$\blacksquare$} $e_s$ (Skill)\\ \textcolor{secondary}{$\blacksquare$} $e_q$ (Quiz)\\ \textcolor{accent}{$\blacksquare$} $e_y$ (Correct)};

    \node[circle, draw=primary, fill=primary!10, thick, minimum size=0.8cm] (h1) at (2, 2.7) {$h_{t-2}$};
    \node[circle, draw=primary, fill=primary!10, thick, minimum size=0.8cm] (h2) at (5, 2.7) {$h_{t-1}$};
    \node[circle, draw=primary, fill=primary!10, thick, minimum size=0.8cm] (h3) at (8, 2.7) {$h_{t}$};
    
    \draw[neuralarrow] (0.8, 2.7) -- (h1);
    \draw[neuralarrow] (h1) -- (h2); 
    \draw[neuralarrow] (h2) -- (h3);
    \foreach \xpos in {2, 5, 8} { \draw[neuralarrow] (\xpos, 1.6) -- (\xpos, 2.3); }

    \node[draw=accent, fill=accent!10, thick, rounded corners=4pt, font=\sffamily\scriptsize] (mast) at (3.5, 4.4) {\texttt{mastered}};
    \node[draw=accent, fill=accent!10, thick, rounded corners=4pt, font=\sffamily\scriptsize] (avg) at (6.5, 4.) {\texttt{avg\_emb}};

    \draw[symbolicarrow, dashed, shorten <=2pt, shorten >=2pt]
      ([xshift=-2pt,yshift= 2pt]x1.north west) to[out=110,in=190,looseness=1.15] (mast.west);
    \draw[symbolicarrow, dashed, shorten <=2pt, shorten >=2pt]
      ([xshift=-2pt,yshift= 2pt]x2.north west) to[out=110,in=250,looseness=1.10] (mast.south);
    \draw[symbolicarrow, dashed, shorten <=2pt, shorten >=2pt]
      ([xshift=-2pt,yshift= 2pt]x3.north west) to[out=110,in=350,looseness=1.10] (mast.south east);

    \draw[symbolicarrow, dashed, shorten <=2pt, shorten >=2pt]
      (2, 1.6) to[out=50,in=185,looseness=1.12] (avg.south west);
    \draw[symbolicarrow, dashed, shorten <=2pt, shorten >=2pt]
      (5, 1.6) to[out=10,in=270,looseness=1.08] (avg.south);
    \draw[symbolicarrow, dashed, shorten <=2pt, shorten >=2pt]
      (8, 1.6) to[out=150,in=315,looseness=1.12] (avg.south east);

    \node[ellipse, draw=accent!80, fill=primary!5, thick, minimum size=0.8cm] (fuse) at (9.0, 3.5) {$joint_t$};
    
    \draw[neuralarrow] (h3) -- (fuse);
    \draw[symbolicarrow, dashed, shorten <=2pt, shorten >=2pt]
      (mast.east) to[out=0,in=150,looseness=1.05] (fuse.north west);
    \draw[symbolicarrow, dashed, shorten <=2pt, shorten >=2pt]
      (avg.east)  to[out=10,in=160,looseness=1.08] (fuse.north west);

    \node[circle, draw=secondary, fill=secondary!10, thick, minimum size=1cm] (out) at (10.6, 3.5) {$\hat{y}_{t+1}$};
    \draw[jointarrow] (fuse) -- (out);
        
    \node[font=\sffamily\tiny, text=primary, anchor=west] at (0.2, 4.8) {$\longrightarrow$ Neural Ops};
    \node[font=\sffamily\tiny, text=accent, anchor=west] at (0.2, 4.4) {$\dashrightarrow$ Symbolic Ops};
\end{scope}

\draw[flowarrow, line width=1.5pt] (16.25, -4.8) -- (16.25, -5.5);

\fill[phasebox] (-1, -13.5) rectangle (21, -21.0);
\node[phasetitle, anchor=north] at (10, -13.5) {4. Model Evaluation \& Interpretation};

\draw[bordergray, thick] (6.5, -14.5) -- (6.5, -20.5);
\draw[bordergray, thick] (13.5, -14.5) -- (13.5, -20.5);

\node[font=\sffamily\bfseries, text=darktext] at (2.75, -14.5) {Predictive Performance};

\begin{scope}[shift={(0.55, -17.35)}, xscale=2.05, yscale=0.85]
    \draw[->, thick, darktext] (0,0) -- (2.5,0) node[below, font=\sffamily\tiny] {FPR};
    \draw[->, thick, darktext] (0,0) -- (0,2.5) node[above, font=\sffamily\tiny] {TPR};
    \fill[primary!20] (0,0) .. controls (0.2,1.8) and (0.8,2.2) .. (2.2,2.2) -- (2.2,0) -- cycle;
    \draw[primary, thick] (0,0) .. controls (0.2,1.8) and (0.8,2.2) .. (2.2,2.2);
    \draw[bordergray, dashed] (0,0) -- (2.2,2.2);
    \node[font=\sffamily\bfseries, text=darktext] at (1.1, 0.95) {AUC 0.90};
    \node[font=\sffamily\scriptsize, align=center] at (1.25, -0.75) {ROC Curve};
\end{scope}

\begin{scope}[shift={(0.55, -20.0)}, xscale=2.65, yscale=0.78]
    \foreach \x in {0, 0.3, 0.6, 0.9, 1.2, 1.5} {
        \foreach \y in {0, 0.3, 0.6, 0.9, 1.2, 1.5, 1.8} {
            \pgfmathsetmacro\col{rnd*100}
            \fill[secondary!\col!white, draw=white, thick] (\x,\y) rectangle (\x+0.3, \y+0.3);
        }
    }
    \draw[->, thick, darktext] (0,0) -- (1.8,0) node[below, font=\sffamily\tiny] {Time};
    \draw[->, thick, darktext] (0,0) -- (0,2.1) node[above, font=\sffamily\tiny] {Skills};
    \node[font=\sffamily\scriptsize, align=center] at (0.9, -0.75) {Mastery Prediction Heatmaps};
\end{scope}

\node[font=\sffamily\bfseries, text=darktext] at (10.0, -14.5) {Computation Graph Logic Tracking};

\begin{scope}[shift={(10.0, -16.0)}]
    \node[draw=secondary, fill=secondary!10, rounded corners=4pt, thick, font=\sffamily\small, minimum width=2.7cm, minimum height=0.9cm, align=center] (target) at (0, 0) {Prediction\\$\hat{y}_{t+1}$};

    \node[draw=primary, fill=primary!10, rounded corners=4pt, thick, font=\sffamily\small, minimum width=2.3cm, minimum height=0.8cm, align=center] (nn) at (-1.6, -1.9) {RNN Path ($h_t$)};
    \node[draw=accent, fill=accent!10, rounded corners=4pt, thick, font=\sffamily\small, minimum width=2.3cm, minimum height=0.8cm, align=center] (rule) at (1.6, -1.9) {Rule Path ($\mathcal{R}$)};

    \node[draw=darktext, fill=white, rounded corners=2pt, font=\sffamily\scriptsize, minimum width=1.8cm] (x1) at (-2.4, -3.8) {Inputs $x_{1..t}$};
    \node[draw=darktext, fill=white, rounded corners=2pt, font=\sffamily\scriptsize, minimum width=1.6cm] (xt) at (0, -3.8) {Latent $z_t$};
    \node[draw=darktext, fill=white, rounded corners=2pt, font=\sffamily\scriptsize, minimum width=1.8cm] (rulesym) at (2.4, -3.8) {\texttt{Mastered(t)}};

    \draw[flowarrow] (nn) -- (target);
    \draw[flowarrow] (rule) -- (target);
    \draw[flowarrow] (x1) -- (nn);
    \draw[flowarrow] (x1) -- (rule);
    \draw[flowarrow] (xt) -- (nn);
    \draw[flowarrow] (rulesym) -- (rule);

    \draw[flowarrow] (xt) .. controls (0.5, -3.0) and (0.8, -2.6) .. (rule.south);

    \draw[->, thick, red, dashed, line width=1.2pt] (rulesym) -- (rule)
        node[midway, right=2pt, yshift=4pt, font=\sffamily\tiny] {\textbf{Activated}};
    \draw[->, thick, red, dashed, line width=1.2pt] (rule) -- (target)
        node[midway, right=2pt, font=\sffamily\tiny] {\textbf{Modulates}};

    \node[font=\sffamily\scriptsize, align=center, text=darktext] at (0, -4.5) {Tracing logical connections to predictions};
\end{scope}

\node[font=\sffamily\bfseries, text=darktext] at (17.25, -14.5) {Attribution-Based Interpretability};

\begin{scope}[shift={(15.0, -19.2)}]
    \draw[->, thick, darktext] (0,0) -- (4.5,0)
        node[below=2pt, font=\sffamily\tiny] {Attribution Score ($\nabla_{\text{input}}$)};
    \draw[thick, darktext] (0,0) -- (0,3.5);

    \fill[accent!80, draw=darktext, thick] (0, 2.6) rectangle (3.8, 3.1);
    \node[font=\sffamily\scriptsize, anchor=west] at (0.12, 2.85) {$R_{mast}$ (Rule)};

    \fill[primary!80, draw=darktext, thick] (0, 1.8) rectangle (2.5, 2.3);
    \node[font=\sffamily\scriptsize, anchor=west] at (0.12, 2.05) {$x_t$ (Current)};

    \fill[accent!80, draw=darktext, thick] (0, 1.0) rectangle (1.8, 1.5);
    \node[font=\sffamily\scriptsize, anchor=west] at (0.12, 1.25) {$R_{avg}$ (Rule)};

    \fill[primary!80, draw=darktext, thick] (0, 0.2) rectangle (0.9, 0.7);
    \node[font=\sffamily\scriptsize, anchor=west] at (0.12, 0.45) {$x_{t-1}$};

    \node[font=\sffamily\scriptsize, align=center, text=darktext] at (2.25, -1.2) {Quantifying impact of features \& rules};
\end{scope}

\draw[flowarrow, line width=1.5pt] (10.5, -13) -- (10.5, -13.5);

\end{tikzpicture}
}

%% file: figure/nesy_dkt_graph.tex
\resizebox{\textwidth}{!}{%
\begin{tikzpicture}[
    x=1.1cm, y=0.9cm,
    >=stealth,
    font=\sffamily\scriptsize,
    house/.style={
        draw=none,
        align=center,
        inner sep=8pt,
        minimum width=5mm,
        minimum height=5mm,
        execute at begin node=\vspace{1.0ex}, 
        path picture={
            \draw[black, thick]
              (path picture bounding box.south west) --
              (path picture bounding box.south east) --
              ($(path picture bounding box.south east)!0.72!(path picture bounding box.north east)$) --
              (path picture bounding box.north) --
              ($(path picture bounding box.south west)!0.72!(path picture bounding box.north west)$) -- cycle;
        }
    },
    rule/.style={
        ellipse,
        draw=red,
        thick,
        align=center,
        inner sep=2pt
    },
    target/.style={
        chamfered rectangle,
        draw=blue!80,
        thick,
        align=center,
        inner sep=4pt
    },
    matrixlabel/.style={
        fill=white,
        align=center,
        inner sep=1pt,
        font=\small\sffamily
    },
    modulebox/.style={
        draw=black!60,
        dashed,
        rounded corners=6pt,
        thick,
        fill=orange!2,
        inner sep=8pt
    },
    modulelabel/.style={
        font=\sffamily\footnotesize\bfseries,
        text=black!80,
        fill=white,
        inner sep=2pt
    }
]


    \node[target] (N41) at (0, 0) {
        \colorbox{yellow!40}{correct(t1, q954)}
    };

    \node[rule] (N28) at (-8, 3) {
        \colorbox{yellow!40}{correct(t1, q954)} :-\\
         final\_nn\_out(t1)
    };

    \node[rule] (N39) at (0, 3) {
        \colorbox{yellow!40}{correct(t1, q954)} :-\\
        avg\_embed(t1, q954)
    };

    \node[rule] (N34) at (8, 3) {
        \colorbox{yellow!40}{correct(t1, q954)} :-\\
        mastered(q954, t1)
    };

    \node[rule] (N_final) at (-8, 5) {
        final\_nn\_out(t1) :-\\
        rnn\_1\_out(t1)
    };

    \node[rule] (N22) at (-8, 7) {
        rnn\_1\_out(t1) :-\\
        combined\_embed(t0), rnn\_1\_out(t0),\\ $next$(t0, t1)
    };

    \node[rule] (N38) at (0, 7) {
        avg\_embed(t1, q954) :-\\
        combined\_embed(t0), quiz\_input(t0, q954),\\ $less$(t0, t1)
    };

    \node[rule] (N31) at (8, 7) {
        mastered(q954, t1) :-\\
        correct\_input(t0, right), quiz\_input(t0, q954),
        \\ $next$(t0, t1)
    };

    \node[rule] (N_init) at (-8, 9) {
        rnn\_1\_out(t0) :-\\
        rnn\_1\_h0
    };

    \node[house] (N18) at (-8, 11) {
        rnn\_1\_h0
    };

    \node[rule] (N15) at (0, 11) {
        combined\_embed(t0) :-\\
        quiz\_embed(t0), skill\_embed(t0), \\ correct\_input\_embed(t0),
    };

    \node[house] (N30) at (8, 11) {
        quiz\_input(t0, q954)
    };

    \node[rule] (N8) at (8, 13.5) {
        correct\_input\_embed(t0) :-\\
        correct\_input(t0, right)
    };

    \node[house] (N0) at (-8, 16) {
        quiz(q954)
    };

    \node[house] (N11) at (0, 16) {
        skill(s3)
    };

    \node[house] (N7) at (8, 16) {
        correct\_input(t0, right)
    };

    \begin{pgfonlayer}{background}
        \node[modulebox,
              fit=(N18) (N_init) (N22) (N_final) (N28),
              label={[modulelabel, xshift=10pt, yshift=25pt, text=red!65]above left:{RNN rules}}] {};

        \node[modulebox,
              fit=(N15) (N38) (N39),
              label={[modulelabel, yshift=10pt, xshift=20pt, text=red!65]above left:{Embedding rules}}] {};

        \node[modulebox,
              fit=(N7) (N30) (N31) (N34),
              label={[modulelabel, xshift=-20pt, yshift=65pt, text=red!65]above right:{Mastery rules}}] {};
    \end{pgfonlayer}


    \draw[->] (N0) -- node[matrixlabel, pos=0.3] {$W_2 \approx 0.90$} (N15);
    \draw[->] (N11) -- node[matrixlabel, pos=0.5] {$W_1 \approx -0.82$} (N15);
    \draw[->] (N8) -- node[matrixlabel, pos=0.4] {$W_3 \approx 0.52$} (N15);

    \draw[->, dashed] (N7) -- node[matrixlabel, pos=0.5] {} (N8);

    \draw[->, dashed] (N7.east) to[out=0, in=0, looseness=1.2] (N31.east);
    \draw[->, dashed] (N0.west) to[out=180, in=180, looseness=1.2] (N28.west);

    \draw[->] (N18) -- node[matrixlabel, pos=0.5] {$W_7 \approx 0.01$} (N_init);
    \draw[->, dashed] (N_init) -- (N22);
    \draw[->] (N15) -- node[matrixlabel, pos=0.6] {$W_6 \approx -0.69$} (N22);
    \draw[->] (N15) -- node[matrixlabel, pos=0.4] {$W_{13} \approx 0.62$} (N38);
    \draw[->, dashed] (N30) -- (N38);
    \draw[->, dashed] (N30) -- node[matrixlabel, pos=0.4] {} (N31);

    \draw[->, dashed] (N22) -- (N_final);
    \draw[->] (N_final) -- node[matrixlabel, pos=0.5] {$W_{14} \approx 0.21$} (N28);
    \draw[->] (N38) -- node[matrixlabel, pos=0.5] {$W_{14} \approx 0.41$} (N39);
    \draw[->] (N31) -- node[matrixlabel, pos=0.5] {$W_{12} \approx 1.09$} (N34);

    \draw[->] (N28) -- node[matrixlabel, pos=0.5] {$W_8 \approx -0.70$} (N41);
    \draw[->] (N39) -- node[matrixlabel, pos=0.5] {$W_10 \approx 0.80$} (N41);
    \draw[->] (N34) -- node[matrixlabel, pos=0.5] {$W_{11} \approx 1.39$} (N41);

\end{tikzpicture}%
}